\documentclass{article}

\usepackage{PRIMEarxiv}

\usepackage[utf8]{inputenc} 
\usepackage{graphicx}
\usepackage{subcaption}
\usepackage{float}
\usepackage[colorlinks=true,linkcolor=blue,citecolor=blue]{hyperref}

\usepackage{soul}
\usepackage{xcolor}

\usepackage{booktabs}
 \usepackage{graphics}
 \usepackage{tabularx}
\usepackage{makecell}
\usepackage[export]{adjustbox}
\usepackage{multirow}
\usepackage{enumitem}
\usepackage{subcaption}
\usepackage{algorithm} 
\usepackage{algpseudocode}
\usepackage[labelfont=bf,textfont={bf}]{caption}
\usepackage{xcolor}
\apptocmd\normalsize{%
 \abovedisplayskip=12pt
 \abovedisplayshortskip=0pt 
 \belowdisplayskip=12pt 
 \belowdisplayshortskip=7pt
}{}{}
\usepackage{physics}
\hypersetup{
    colorlinks=true,                
    breaklinks=true,                
    urlcolor= cyan,                
    linkcolor= red,                     
    bookmarksopen=false,
    filecolor=black,
    citecolor=blue,
    linkbordercolor=pink
}

\renewcommand{\mathbf}{\boldsymbol}

\usepackage{threeparttable}

\usepackage{rotating}
\usepackage{amsmath}
\usepackage{nccmath, mathtools}
\usepackage{svg}
\usepackage{amssymb} 
\usepackage{etoolbox}
\makeatletter
\newcommand*{\rom}[1]{\expandafter\@slowromancap\romannumeral #1@}
\makeatother

\newcount\Comments  
\usepackage{color}
\definecolor{darkgreen}{rgb}{0,0.5,0}
\definecolor{purple}{rgb}{1,0,1}
\newcommand{\kibitz}[2]{\ifnum\Comments=0\textcolor{#1}{#2}\fi}

\newcommand{\edit}[1]{\kibitz{black}      { #1}}
\newcommand{\editnew}[1]{\kibitz{black}      { #1}}

\usepackage{tikz}

\title{Identifying built environment factors influencing driver yielding behavior at unsignalized intersections: A naturalistic open-source dataset collected in Minnesota
\thanks{\textit{\underline{Citation}}: 
\textbf{Li, et al. Identifying built environment factors influencing driver yielding behavior at unsignalized intersections: A naturalistic open-source dataset collected in Minnesota.}} 
}

\author{
  Tianyi Li \\
  Department of Civil Engineering \\
  Saint Louis University \\
  3450 Lindell Blvd, Saint Louis, MO 63103, USA.\\
  \texttt{tianyi.li.1@slu.edu} \\
   \And
  Joshua Klavins \\
  Department of Civil, Environmental, and Geo-Engineering \\
  University of Minnesota \\
  500 Pillsbury Dr. SE, Minneapolis, MN 55455, USA.\\
  \texttt{klavi008@umn.edu} \\
     \And
 Te Xu \\
  Department of Civil, Environmental, and Geo-Engineering \\
  University of Minnesota \\
  500 Pillsbury Dr. SE, Minneapolis, MN 55455, USA.\\
  \texttt{te000002@umn.edu} \\
     \And
  Niaz Mahmud Zafri\\
  Department of Urban and Regional Planning\\
  Bangladesh University of Engineering and Technology\\
  \texttt{zafri@urp.buet.ac.bd} \\
  \AND
  Raphael Stern \\
  Department of Civil, Environmental, and Geo-Engineering \\
  University of Minnesota \\
  500 Pillsbury Dr. SE, Minneapolis, MN 55455, USA.\\
  \texttt{rstern@umn.edu} \\
}

\begin{document}
\maketitle

\begin{abstract}
\edit{Many factors influence the yielding result of a driver-pedestrian interaction, including traffic volume, vehicle speed, roadway characteristics, etc. While individual aspects of these interactions have been explored, comprehensive, naturalistic studies, particularly those considering the built environment's influence on driver-yielding behavior, are lacking. To address this gap, our study introduces an extensive open-source dataset, compiled from video data at 18 unsignalized intersections across Minnesota. Documenting more than 3000 interactions, this dataset provides a detailed view of driver-pedestrian interactions and over 50 distinct contextual variables. The data, which covers individual driver-pedestrian interactions and contextual factors, is made publicly available at \url{https://github.com/tianyi17/pedestrian_yielding_data_MN}.}

Using logistic regression, we developed a classification model that predicts driver yielding based on the identified variables. Our analysis indicates that vehicle speed, the presence of parking lots, proximity to parks or schools, and the width of major road crossings significantly influence driver yielding at unsignalized intersections. Through our findings and by publishing one of the most comprehensive driver-pedestrian datasets in the United States, our study will support communities across Minnesota and the United States in their ongoing efforts to improve road safety for pedestrians and be helpful for automated vehicle design.
\end{abstract}


\section{Introduction}
\renewcommand{\thefootnote}{\arabic{footnote}}

{Data from the National Highway Traffic Safety Administration shows that there were a total of 7,388 pedestrian fatalities in the United States in 2021~\cite{NHTSA}, a 12.5\% increase from the 6,565 pedestrian fatalities in 2020 and the highest number since 1981.} This highlights the pressing issue of pedestrian safety in transportation systems, particularly at intersections where driver-pedestrian interactions are common. While changes to roadways and crosswalks are crucial, addressing the cultural norms that allow for dangerous driver behaviors is equally important~\cite{craig2019pedestrian}. 

At such an interaction, a pedestrian arrives at a marked or unmarked crosswalk at roughly the same time that a driver is approaching the intersection. The outcome of such an interaction can lead to the driver yielding to the pedestrian or vice versa. The lower compliance rate of drivers yielding to pedestrians at uncontrolled crosswalks compared with controlled crosswalks has been documented~\cite{van2001advance,mitman2010driver,morris2020effective}, thus highlighting the significant risk that such intersections pose to pedestrians. However, some previous efforts to understand this interaction often focused on pedestrian fault instead of identifying factors that lead to higher driver non-yielding rates~\cite{sun2003modeling}. 

Understanding the factors that contribute to driver yielding is crucial for traffic safety analysis and pedestrian safety~\cite{zafri2020factors, zafri2022effect, zafri2023walk}. Driver yielding occurs when a driver slows or stops to allow pedestrians to traverse an intersection safely. Simultaneously, a driver-pedestrian interaction is the outcome (yielding or non-yielding of the driver) of a pedestrian arriving at an intersection while a driver is approaching. Therefore, the approaching driver can yield to the pedestrian by either stopping or slowing to accommodate them; alternatively, the driver can decide not to yield, resulting in a non-yielding event. The driver yielding rate (or simply yielding rate) refers to the percentage of drivers who yield to a pedestrian. Addressing the gap in understanding driver non-yielding behavior, our research uses video data to understand how pedestrians interact with the vehicle and built environment and to identify intersection features that lead to higher driver-yielding rates. 

This study focuses on identifying factors that contribute to higher-yielding rates. We present a comprehensive dataset of naturalistic video data collected from unsignalized intersections in Minnesota, capturing driver-yielding behavior in 2021. This analysis is conducted using a machine learning approach, specifically logistic regression. It should be noted that in our preliminary work~\cite{li2021leveraging}, we presented initial findings from three sites, demonstrating the potential value of such data. This manuscript expands upon that work, including data from a total of 18 locations. The main contributions of our study are as follows:

\begin{itemize}

\item {We collect and publish naturalistic driver-pedestrian-yielding data from 18 unsignalized intersections across Minnesota.} To the best of our knowledge, this is one of the most comprehensive driver-pedestrian-yielding data publically available in the US.

\item {We model the outcome of a driver-pedestrian interaction and identify variables that significantly influence the driver's yielding behavior.}
\end{itemize}

The remainder of this article is organized as follows. First, we review relevant literature on driver-yielding behavior. {Next, we present the experimental data collection procedure and summarize the collected data, and this data is also published in coordination with this manuscript, which is then used for statistical analyses.} Finally, we discuss the results of this analysis and conclude with insights into how our models capture the nuances of driver-yielding behavior.

\section{Review of relevant literature}\label{sec:lit}
The section is divided into four main sections: the first identifies driver characteristics, the second discusses pedestrian characteristics and behavior, the third focuses on traffic and vehicle characteristics, and the fourth addresses intersection characteristics related factors influencing driver yielding behavior towards pedestrians, as identified in previous literature. After these four sections, we include another section where we mention the gap that our study aims to address.

\subsection{Driver characteristics}
A number of studies have focused on driver characteristics related factors influencing driver behavior at intersections, encompassing both signalized and unsignalized types, as well as mid-block crossings~\cite{fitzpatrick2006improving, zhao2020modeling, schneider2017evaluation, zafri2022effect}. This research has explored the role of demographic factors such as age and educational background on driver-yielding behavior. \cite{hirun2016factors} observed that older drivers are more prone to yield to pedestrians than their younger counterparts. The educational level of drivers also plays a significant role, with those holding a bachelor's degree demonstrating a higher tendency to yield compared to drivers with lower educational levels~\cite{hirun2016factors}. This trend suggests that awareness and comprehension of traffic regulations, including right-of-way rules and the concept of unmarked crosswalks, are instrumental in guiding driver behavior~\cite{mitman2010driver}.

Further emphasizing the influence of awareness campaigns, \cite{schneider2015pedestrian} noted the effectiveness of educational campaigns in positively shaping driver-yielding practices. A notable example of this is a study by~\cite{zhang2013educational}, which reported a significant decrease in non-yielding behavior, from 61.42\% to 55.19\%, following a targeted campaign on a university campus.

One significant finding is that driver-yielding behavior is further influenced by the presence of a following vehicle~\cite{schroeder2011event, akin2010neural} and the actions of drivers in the opposite direction~\cite{sun2003modeling, schroeder2011event}, highlighting the complexity of factors at play.

\subsection{Pedestrian characteristics and behavior}
In addition to driver characteristics, pedestrian characteristics and behavior at intersections can influence the drivers' yielding behavior~\cite{pawar2016critical,muley2017pedestrians,coughenour2020estimated,zafri2022effect}. For example, there is a methodology for analyzing unsignalized intersections based on gap acceptance behavior~\cite{HCM2010}, and gap acceptance theory is widely used to evaluate the safe operation of the road at uncontrolled intersections~\cite{pawar2016critical}. Pedestrians have to wait for safe gaps to cross at intersections. This is a high risk for pedestrians because the wrong judgment of the accepted gap can result in traffic crashes. From the perspective of a pedestrian, a pedestrian will either accept or reject the gap in the situation of vehicle-pedestrian interaction, and this will force the driver to react differently if the pedestrian accepts a shorter gap.

Studies have shown that drivers are more inclined to yield to older pedestrians than younger ones~\cite{dileep2016study}. A study by~\cite{zafri2022effect} shows that drivers demonstrated a higher tendency to yield to pedestrians who were female, crossing in a group, carrying baggage, not using a mobile phone, making some hand gesture to the driver, and employing the rolling gap strategy while crossing. In addition to that pedestrian assertiveness, such as briskly walking towards the crosswalk, is also a key factor leading to higher-yielding rates~\cite{schroeder2011event, dileep2016study, schneider2018exploratory}. 

Interestingly, visible cues significantly influence driver behavior. Research indicates that pedestrians using a white cane or donning bright clothing are more likely to elicit yielding from drivers, in contrast to those holding an umbrella or wearing less conspicuous attire~\cite{bourquin2011conditions, salamati2013event}. Additionally, the positioning of pedestrians at the crosswalk plays a critical role; those positioned less than one foot from the curb are observed to have a higher likelihood of being yielded to~\cite{geruschat2005driver}.

\editnew{Drivers are less likely to yield to pedestrians outside marked crosswalks, and they tend to decelerate more for pedestrians within a crosswalk than for those outside it. Collectively, these insights contribute to a nuanced understanding of pedestrian dynamics at intersections, highlighting areas for potential intervention and planning.}

\subsection{Vehicle and traffic characteristics}
The influence of vehicle characteristics on driver-yielding behavior is a subject of considerable research interest. A notable finding is the tendency of drivers in high-status, expensive vehicles in the United States to yield less frequently to pedestrians, indicating a relationship between vehicle status and driver behavior~\cite{piff2012higher}.

In addition to vehicle status, speed is a critical factor that significantly impacts driver yielding. Research has consistently shown that lower vehicle speeds are associated with higher yielding rates~\cite{zheng2015modeling, schroeder2011event, salamati2013event, silvano2016analysis, theofilatos2021cross}. \cite{geruschat2005driver} observed this trend in a study focusing on a two-lane roundabout. They reported that a decrease in vehicle speed to 15 miles per hour resulted in a yielding rate increase of up to 75\%. In contrast, an increment of merely five miles per hour led to a notable reduction in yielding rates, dropping them to about 50\%. \editnew{Furthermore, specific vehicle characteristics, such as model and type, also play a significant role in influencing driver behavior at intersections~\cite{akin2010neural}.}

Beyond individual vehicle attributes, the overall traffic dynamics, such as vehicular volume and the presence of vehicle platoons, crucially affect driver-yielding behavior. Studies have specifically highlighted the impact of vehicle platoons, showing that they tend to decrease the likelihood of drivers yielding to pedestrians~\cite{schroeder2011event, alhajyaseen2012estimation}.

\subsection{Intersection characteristics}

The design and layout of intersections play a pivotal role in driver behavior towards pedestrians. This aspect of road safety, extending beyond individual driver or pedestrian behavior, has been the focus of various studies. The idea is that certain intersection characteristics may enhance or compromise pedestrian safety, influencing how welcoming these spaces are for pedestrians.

Research has consistently shown that the physical layout of roads and their surrounding environments significantly impacts driver-yielding behavior. One key finding is that drivers tend to yield less frequently on roads where pedestrians face longer crossing distances~\cite{fitzpatrick2014characteristics, sanatizadeh2017contributing}. \cite{schneider2018exploratory} reported lower yielding rates on major roadways, such as arterial and collector, particularly at unmarked crosswalks, compared to local roads with marked crosswalks.

The angle of intersection is another crucial factor. Studies indicate that left-turning drivers' yielding behavior is influenced by the intersection angle. Drivers at obtuse-angled intersections tend to exhibit more non-yielding behavior, whereas those at acute-angled intersections are more likely to yield~\cite{iasmin2015yielding,alhajyaseen2012estimation}. Moreover, the geometric characteristics of intersections also affect vehicle speed and the potential for collisions.

Effective engineering treatments, including signalization, crossing marks, red signal or beacon treatments, and pedestrian-activated signals, have been shown to significantly influence yielding behavior~\cite{turner2006motorist,hourdos2020assessing}. Additionally, road signs prompting drivers to yield and pedestrian lights, though their activation rates may vary with traffic conditions, have been effective in increasing safe yielding distances~\cite{van1992influence}.

Further emphasizing the importance of design, features such as marked pedestrian crossings, high-visibility crosswalks, overhead warning signs, rectangular rapid flashing beacons, advance yield signs, and distinct road markings have been identified as effective in enhancing pedestrian safety. These elements encourage drivers to yield more readily and are critical in guiding the construction of new intersections or the modification of existing ones to improve pedestrian safety~\cite{schneider2015pedestrian, sanatizadeh2017contributing}.

\subsection{Study gap}

Based on the literature reviewed, several features related to drivers, pedestrians, traffic, vehicles, and intersection characteristics lead to higher driver-yielding rates. However, previous research does not consider the effects of the surrounding built environment on driver-yielding behavior towards pedestrians. Numerous built environment factors, such as the presence of parking, bike lanes, schools, varying land use types, and bus stops, have been recognized for their influence on pedestrian crashes and their severity level~\cite{xin2017effects, prato2018considering, zahabi2011estimating,aziz2013exploring}. Despite these built environmental elements having the potential to significantly affect how drivers respond to pedestrians, detailed investigations into the relationship between these site-specific built environment features and driver yielding are sparse. This research gap is what our study aims to address, utilizing large-scale open-source data to explore the potential influence of built environment factors on driver-yielding behavior in a comprehensive manner.

\section{Data Collection and Description}
One of the contributions of this work is the collection of a large-scale, open-source naturalistic dataset for driver yielding. We collected over 3,300 driver-pedestrian interactions at 18 different unsignalized intersections. The dataset, which focuses on driver-pedestrian interactions, is unique in both its scale and the depth of information it provides. In this section, we first describe site characteristics and the data collection technique, followed by a summary of the selected sites, and finally present summary statistics of the collected data. It is important to note that the sites are anonymized in our reporting to maintain privacy standards.

\subsection{Site Characteristics}
The selection of sites was guided by a set of criteria derived from the literature reviewed in Section~\ref{sec:lit}. These criteria include factors like traffic volume, pedestrian density, and urban setting, as identified by Minnesota Department of Transportation (MnDOT) staff. Our chosen sites, spread across diverse geographical locations, provide a representative cross-section of various urban traffic scenarios and were selected with the guidance of MnDOT staff.

The table~\ref{tab:sites} below summarizes the characteristics of each site, including the number of lanes, posted speed limits, type of road markings, AADT, and intersection geometry. This diversity in site characteristics allows for a robust analysis of driver-pedestrian interactions under different traffic conditions.

\begin{table}[h]

\centering
\begin{adjustbox}{scale=0.83}
\begin{threeparttable}

\caption{Site Characteristics.}
\begin{tabular}{ccccccc}
\hline
\textbf{Site} & \textbf{\makecell{Num. of Lanes \\ (Main Street)}} & \textbf{\makecell{Num. of Lanes \\ (Minor Street)}} & \textbf{Posted Speed (MPH)} & \textbf{Markings}  & \textbf{AADT} & \textbf{Shape} \\
\hline
\textbf{Site 1}  & 4 & 2 &  35  &  Unmarked  & 14600 & T-shape \\
\textbf{Site 2}  & 2 & 2 &  35  &  Unmarked  & 14800 & T-shape \\
\textbf{Site 3}  & 2 & 1 &  30  &  Standard\tnote{1}  & 14800 & T-shape \\
\textbf{Site 4}  & 2 & 1 &  30  &  Unmarked  & 10700 & four-way \\
\textbf{Site 5}  & 2 & 1 &  30  &  Unmarked  & 10500 & four-way \\
\textbf{Site 6}  & 3 & 1 &  35  & Continental  & 6200 & four-way \\
\textbf{Site 7}  & 4 & 1 &  35  & Continental  & 3300 & T-shape \\
\textbf{Site 8}  & 4 & 1 &  30  &  Unmarked  & 10200 & four-way \\
\textbf{Site 9}  & 3 & 1 &  35  &  Unmarked  & 5400 & four-way \\
\textbf{Site 10} & 4 & 1 &  35  &  Unmarked  & 18800 & four-way \\
\textbf{Site 11} & 3 & 1 &  30  &  Unmarked  & 18100 & four-way \\
\textbf{Site 12} & 2 & 1 &  30  &  Unmarked  & 10300 & four-way \\
\textbf{Site 13} & 2 & 1 &  35  & Continental & 7400 & four-way \\
\textbf{Site 14} & 2 & 1 &  30  &  Standard  & 7100  & four-way \\
\textbf{Site 15} & 3 & 1 &  30  &  Unmarked  & 16400 & four-way \\
\textbf{Site 16} & 2 & 1 &  30  &  Standard  & 12100 & four-way \\
\textbf{Site 17} & 4 & 1 &  30  &  Unmarked  & 12800 & T-shape \\
\textbf{Site 18} & 4 & 1 &  30  &  Unmarked  & 12500 & four-way \\
\hline
\label{tab:sites}
\end{tabular}

\begin{tablenotes}
\item[1] Standard markings refer to markings with two solid white lines.
\end{tablenotes}
\end{threeparttable}
\end{adjustbox}

\end{table}



\subsection{Data Collection and Cleaning}
\begin{figure}
    \centering
    \includegraphics[scale=0.15]{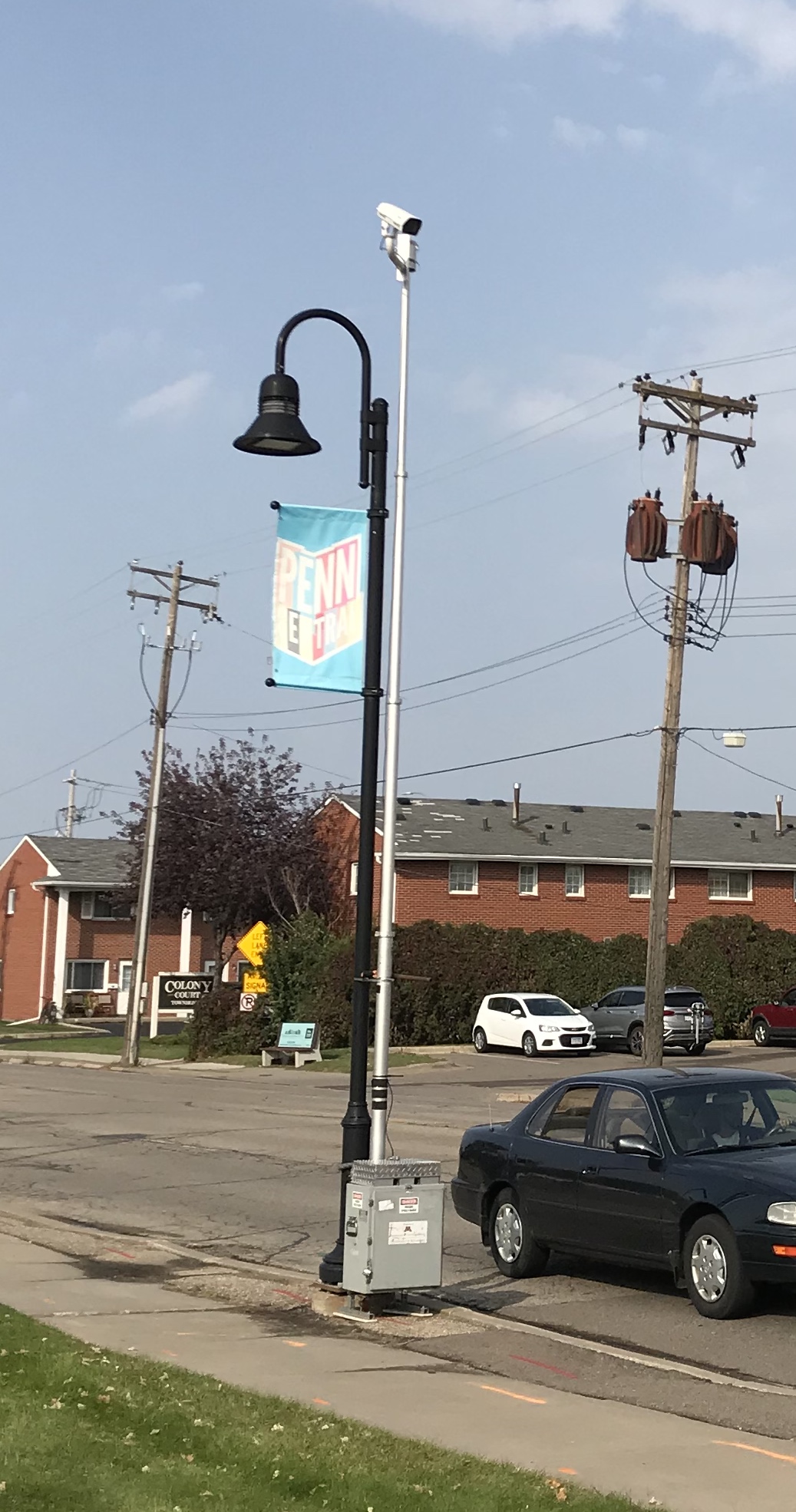}
    \caption{{TIM sensor attached to a light pole during data collection at site 17.}}
    \label{fig:camera}
\end{figure}

To investigate driver-yielding behavior, we utilized the traffic information monitor (TIM) platform developed by the Minnesota Traffic Observatory (MTO) at the University of Minnesota. The TIM is a customized sensor pack that consists of a pole-mounted video camera that can extend up to 30 feet, 12 V batteries that can sustain up to four weeks of operation, a Raspberry Pi computer, and a watertight metal enclosure to protect the batteries and computer. In this study, we used the TIM and the camera to capture video data of pedestrians and drivers approaching intersections in a naturalistic setting, as depicted in Fig.~\ref{fig:camera}.

In the summer of 2021, we strategically deployed TIMs across 18 diverse sites in Minnesota. These sites were chosen to encapsulate a broad spectrum of pedestrian and driver interactions, environmental contexts, and traffic patterns. The TIMs recorded extensive video data during daylight hours for one to two weeks at each site, focusing on capturing naturalistic driver and pedestrian behavior at intersections.

We collected more than 50 features for each interaction, and each pair of interactions needs to be collected individually. To extract the driver-pedestrian interactions accurately and analyze the data efficiently, a trained data collector manually extracted individual interaction events of all the variables. The speed extraction methodology computes the vehicular speed of the concerned driver-pedestrian interaction by measuring the time each vehicle takes to pass two known positions at the intersection. Building upon preliminary work cited in~\cite{li2021leveraging}, which involved data from three sites, this study expands the dataset to include 18 locations. 


\edit{In this study, a ``codable driver-pedestrian interaction" is characterized as an event where either the driver yields to the pedestrian or the pedestrian forces to yield for giving the right of way to the vehicle. A ``close call'' is defined as a near-miss incident in which an imminent collision between a vehicle and a pedestrian is narrowly averted, often necessitating quick evasive actions by either or both parties involved. It's worth noting that this measure is somewhat subjective, especially given that data collection was performed by a single individual. The term ``number of bus stops'' refers to the count of bus stops located within one block of the main road intersecting the observed location. For further details on the variables collected and their implications, we direct readers to our publicly available repository of all the collected data and Tables~\ref{tab:sites} and~\ref{tab:stat_locations} in the following sections, and Table~\ref{tab:Variables_description} in the Appendix. We acknowledge the potential subjectivity in data collection. To mitigate this, we provided a clear definition of the variables during the data collection procedure. Additionally, having one consistent data collector, who worked on this project for over a year, helped ensure uniformity in the data collection process.}

\subsection{Data Description}
\editnew{Table~\ref{tab:stat_locations} presents the summary statistics for pedestrian behavior at the 18 intersections studied. Factors include the number of total interactions, the number of close calls, and the driver's yielding rate. Intersection 16 had the most interactions with 840 events, and intersection 11 had the fewest number of interactions with 20. While intersection 16 also had the highest yielding rate of 70.36\%, intersection 11 had the lowest yielding rate of 0. This suggests that locations with higher pedestrian volumes tend to have higher driver-yielding rates. Overall, we found that yielding rates are generally low, as shown in Table~\ref{tab:stat_locations}.} For example, at Site 5, the yielding rate was particularly low, and at intersections 11 and 15, the yielding rate was 0, indicating that no drivers yielded to pedestrians during our study period. This is likely due to the environmental factors at these sites, where drivers may not anticipate yielding to pedestrians. Another noteworthy observation is that the maximum waiting time for pedestrians at certain intersections approaches 1 minute. In some instances, pedestrians were forced to halt midway through crossing, potentially elevating the risk of crashes. The diminished yielding rate at unsignalized, unmarked intersections with low pedestrian volume may be attributed to drivers' lack of anticipation for pedestrian presence.
\begin{figure}
    \centering
    \begin{subfigure}{\textwidth}
        \centering
        \includegraphics[scale=0.7]{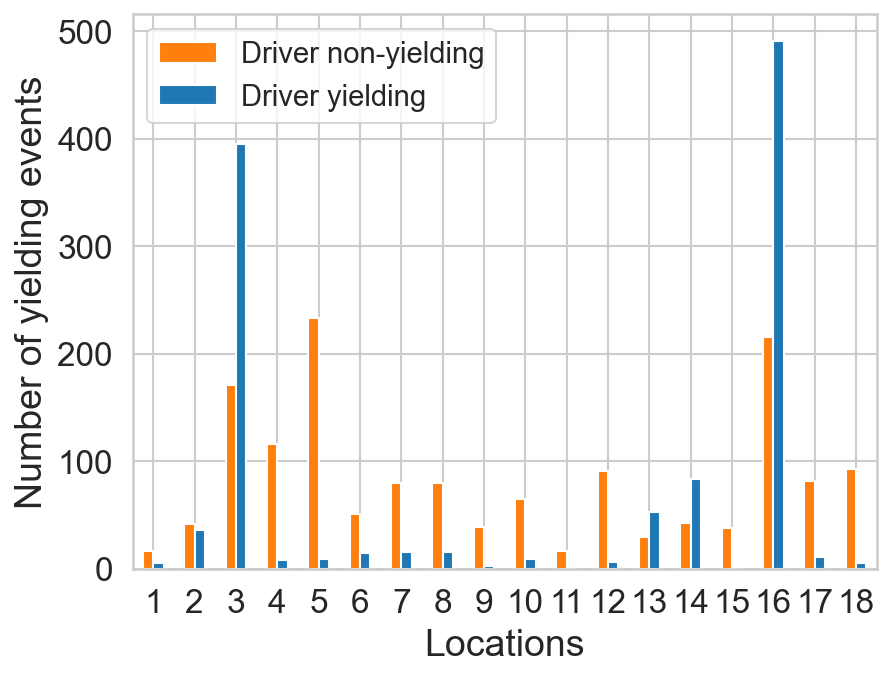}
        \caption{Location ID vs. the number of events.}
        \label{fig:sub_locations}
    \end{subfigure}
    
    \begin{subfigure}{\textwidth}
        \centering
        \includegraphics[scale=0.7]{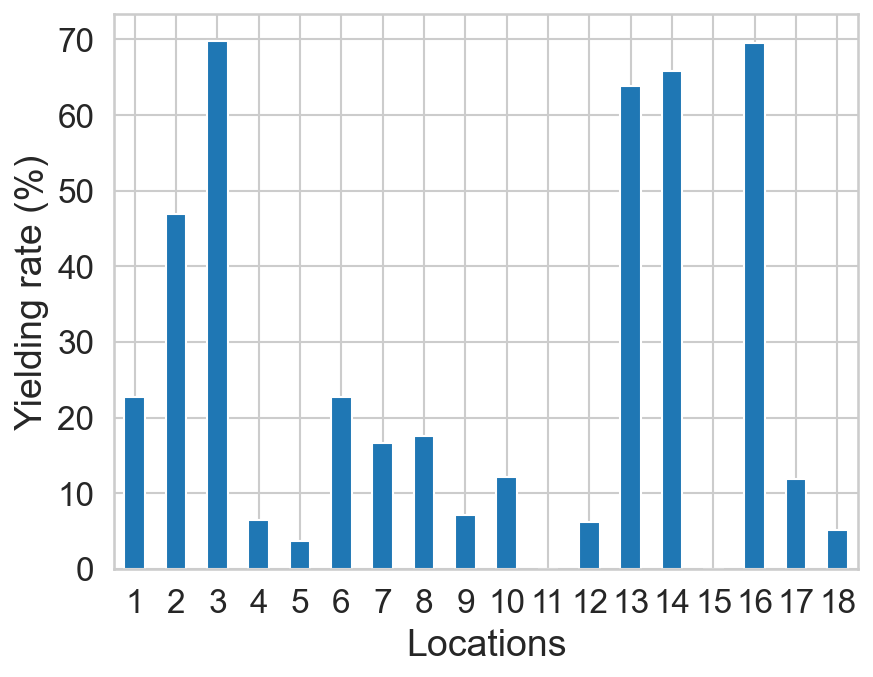}
        \caption{Location ID vs. yielding rate.}
        \label{fig:sub_locations_rate}
    \end{subfigure}
    
    \caption{Location ID vs. yielding outcome showing different yielding rates among sites.}
    \label{fig:locations}
     \end{figure}

\begin{figure}
    \centering 
\begin{subfigure}{0.49\textwidth}
\includegraphics[width=\linewidth]{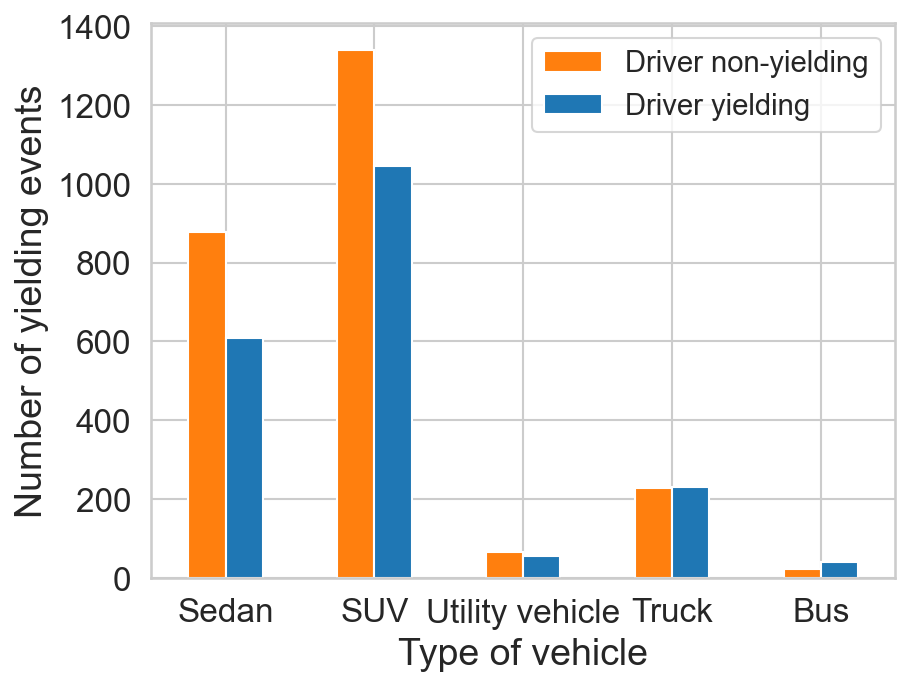}
\caption{Type of vehicles vs. yielding. \label{fig:vehicles}}
\end{subfigure}\hfil 
\medskip
\begin{subfigure}{0.49\textwidth}
\includegraphics[width=\linewidth]{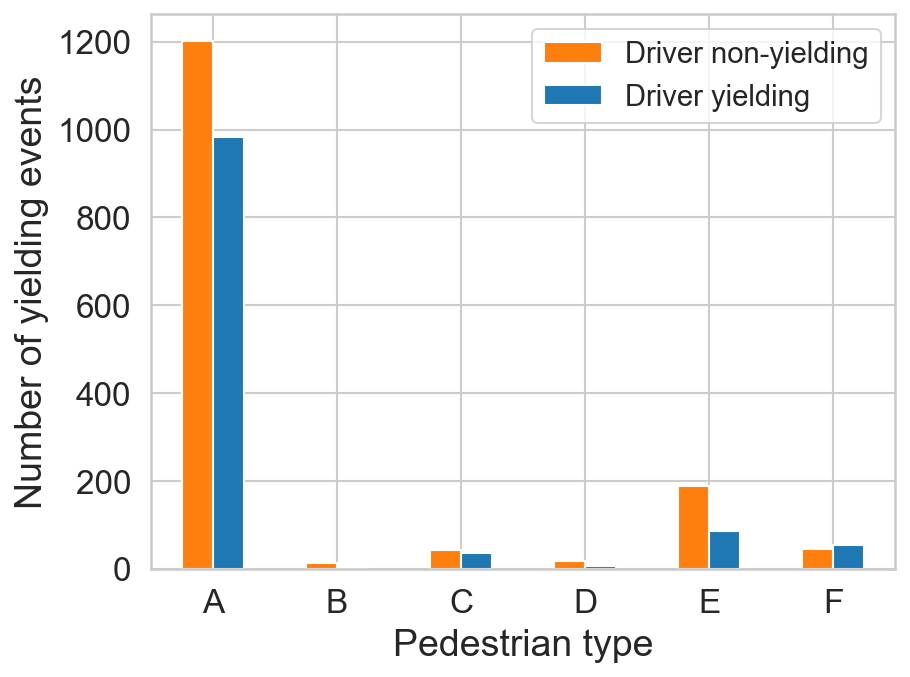}
\caption{Type of pedestrian vs. yielding. \label{fig:ped}}
\end{subfigure}\hfil 
\begin{subfigure}{0.49\textwidth}
\includegraphics[width=\linewidth]{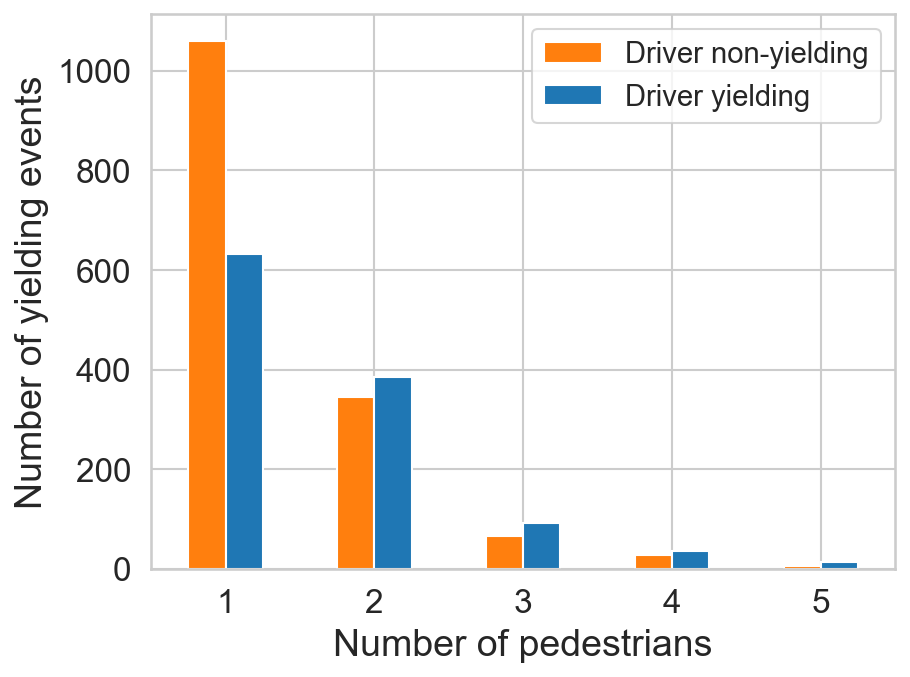}
\caption{Number of pedestrians vs. yielding. \label{fig:ped_num}}
\end{subfigure}\hfil 
\begin{subfigure}{0.49\textwidth}
\includegraphics[width=\linewidth]{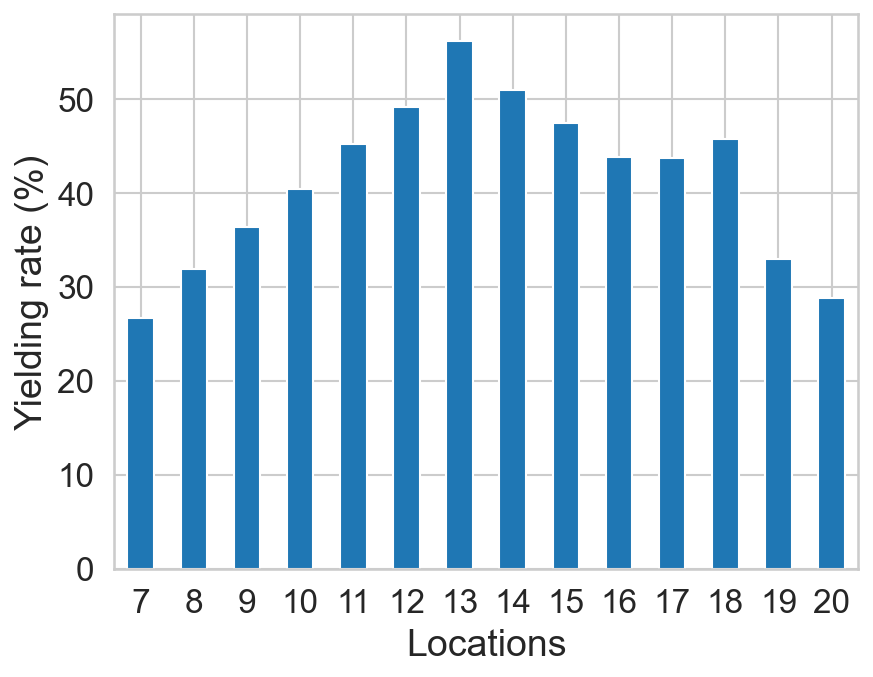}
\caption{Hour of the day vs. yielding rate. \label{fig:hours}}
\end{subfigure}\hfil 
\caption{Visualization of predictor variables: Event Features}
\label{fig:vis_1}
\end{figure}

Fig.~\ref{fig:vehicles} demonstrates the influence of vehicle type on yielding rates. The results indicate that Sports Utility vehicles (SUVs) and sedans are less likely to yield compared to other vehicle types, whereas buses and trucks exhibit the highest yielding rates.
We classified pedestrians into six distinct categories to examine differences in driver interactions, as illustrated in Fig.~\ref{fig:ped}: 
A: \textit{pedestrian}, B: \textit{person on a vehicle such as a scooter}, C: \textit{mix of pedestrian types: scenarios involving more than one pedestrian}, D: \textit{others}, E: \textit{person with a dog}, and F: \textit{person with a stroller or child}. The most common type is Category A and the second most common one is pedestrian with a dog (Category E). As exhibited by Fig.~\ref{fig:ped_num}, the drivers are more likely to yield to a group of pedestrians and less likely to yield to a pedestrian alone. In this study, interactions between cyclists and vehicles have been excluded, though the interactions are kept in the published dataset for completion. Fig.~\ref{fig:hours} illustrates the variations in yielding behavior (yielding rate) at different hours of the day. Our findings indicate that drivers are more likely to yield during noon compared to other times throughout the day.

\begin{figure}
    \centering 
\begin{subfigure}{0.49\textwidth}
\includegraphics[width=\linewidth]{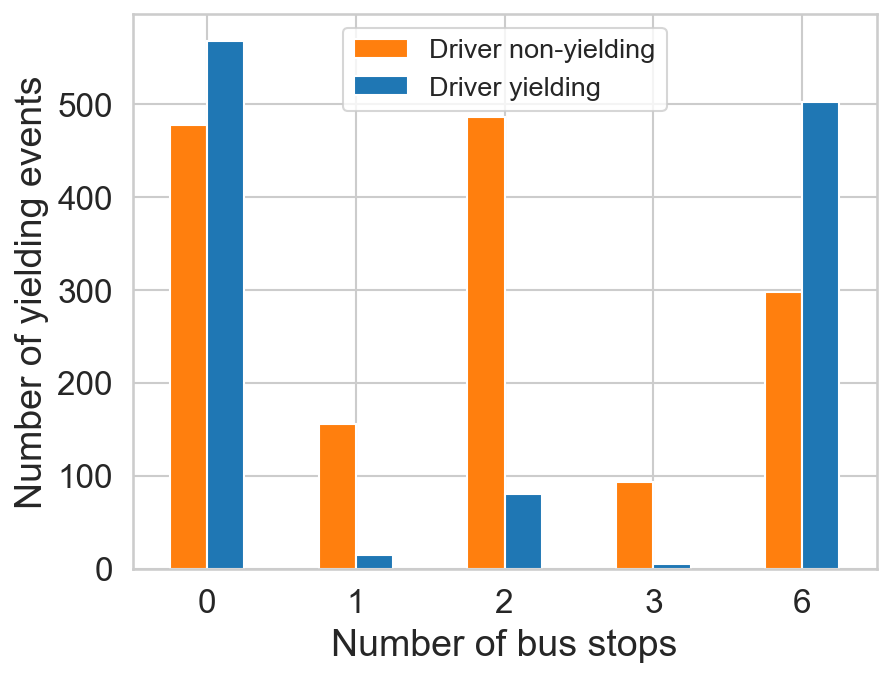}
\caption{Number of bus stops vs. yielding. \label{fig:bus}}
\end{subfigure}\hfil 
\medskip
\begin{subfigure}{0.49\textwidth}
\includegraphics[width=\linewidth]{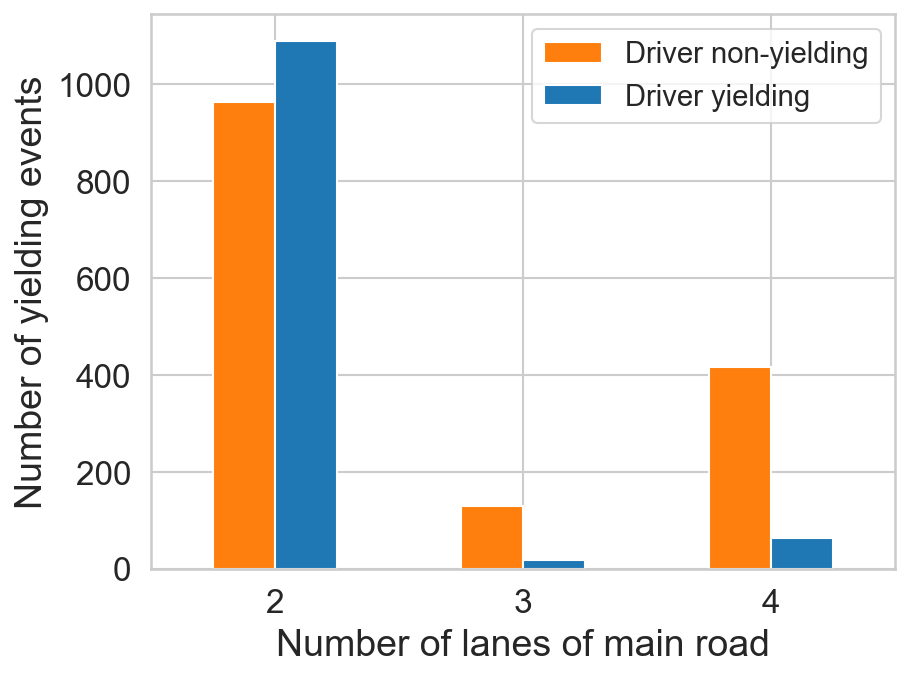}
\caption{Number of lanes vs. yielding. \label{fig:lanes}}
\end{subfigure}\hfil 
\begin{subfigure}{0.49\textwidth}
\includegraphics[width=\linewidth]{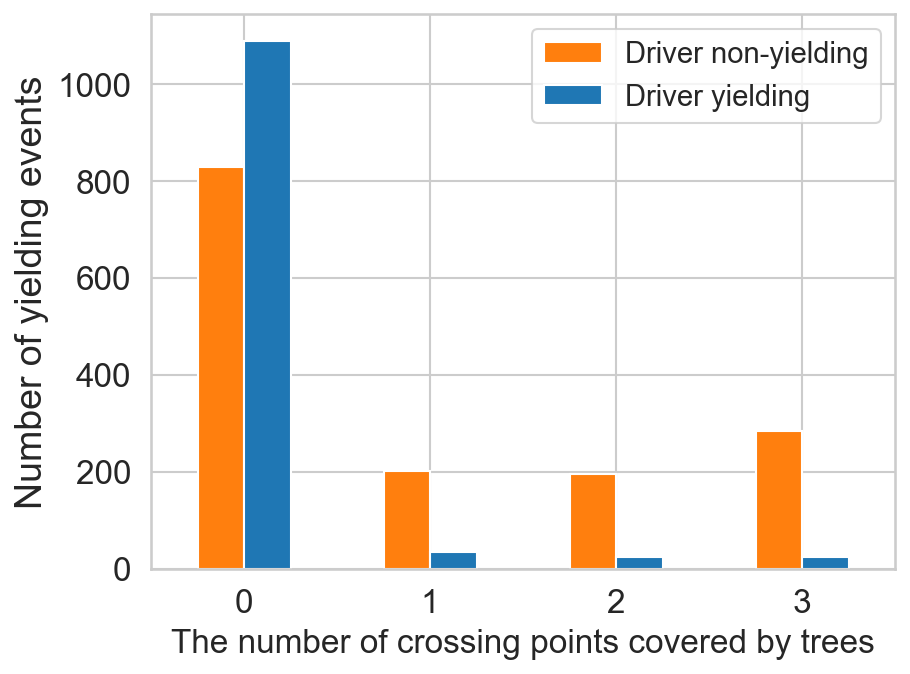}
\caption{Tree cover vs. yielding. \label{fig:tree}}
\end{subfigure}\hfil 
\caption{Visualization of predictor variables: Pedestrian and vehicle characteristics}
\label{fig:vis_2}
\end{figure}

{We collected variables that related to the site location to reflect location characteristics so that the results can be used to improve pedestrian walkability. 
For example, as illustrated in Fig.~\ref{fig:bus}, we observed a correlation between the number of bus stops near an intersection and driver yielding rates, for the intersection with bus stops, the more bus stops, the more drivers yield to pedestrian.} Notably, intersections with a higher number of bus stops exhibited increased rates of driver yielding. This finding suggests that the presence of bus stops may influence driver awareness and behavior toward pedestrians. Interestingly, this correlation was not observed at intersections without nearby bus stops, indicating a unique relationship between bus stop density and yielding behavior. Regarding the number of lanes on the major road (Fig.~\ref{fig:lanes}), drivers are less likely to yield when there are more lanes. Fig.~\ref{fig:tree} reveals a correlation between lower driver-yielding rates and the presence of trees obstructing the drivers' view of waiting pedestrians. \editnew{Specifically, the results indicate that the highest driver-yielding rates occur at intersections without any trees blocking the view of crosswalk, while the lowest yielding rate is found at intersections where tree cover is present at most of the intersection corners.}

\begin{figure}
\centering 
\begin{subfigure}{0.49\textwidth}
\includegraphics[width=\linewidth]{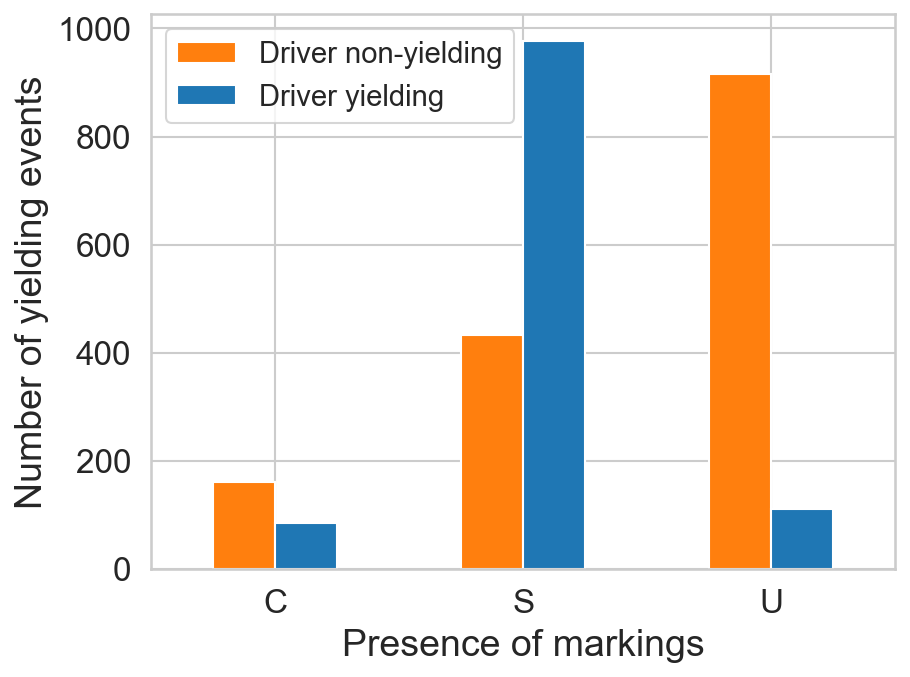}
\caption{Comparison of Driver Yielding Rates by Crosswalk Marking Types. 'C' represents continental markings, 'S' denotes standard markings with two parallel lines, and 'U' indicates unmarked crosswalks. \label{fig:markings}}
\end{subfigure}\hfil 
\begin{subfigure}{0.49\textwidth}
\includegraphics[width=\linewidth]{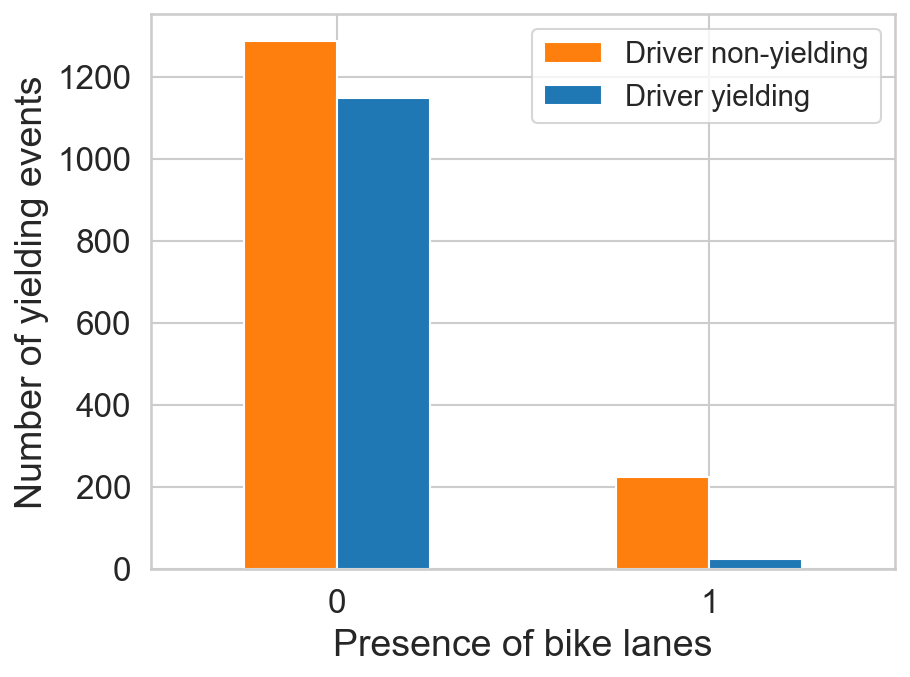}
\caption{Impact of Bike Lane Presence on Driver Yielding. '0' denotes the absence of bike lanes, while '1' indicates their presence. \label{fig:bike_lane}}
\end{subfigure}\hfil 
\medskip
\begin{subfigure}{0.49\textwidth}
\includegraphics[width=\linewidth]{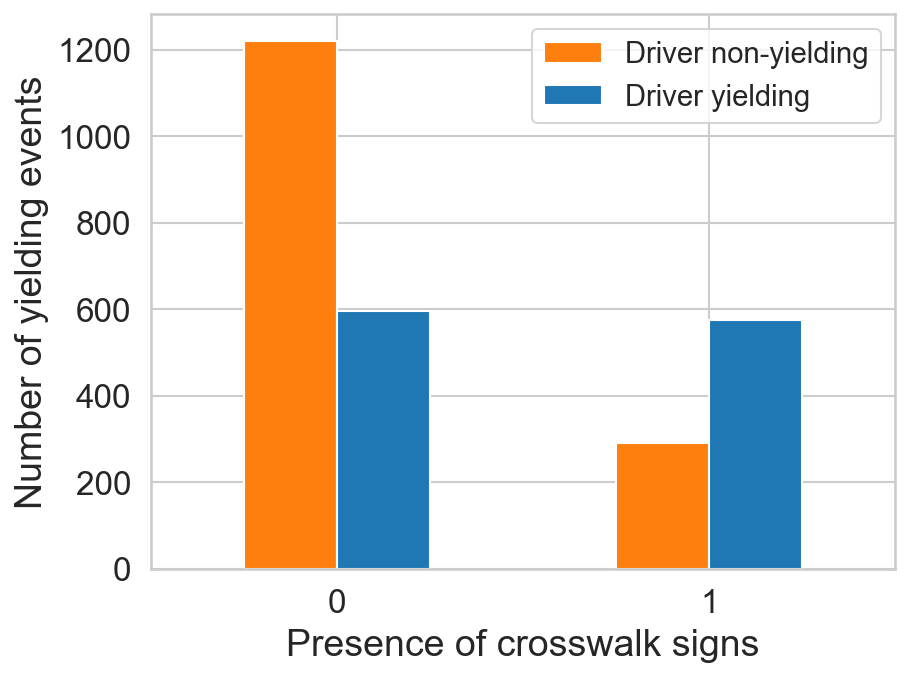}
\caption{Relationship between crosswalk sign presence and driver yielding. '0' represents no crosswalk signs, and '1' signifies the presence of crosswalk signs. \label{fig:signage}}
\end{subfigure}\hfil 
\caption{Visualization of predictor variables: Signs and markings}
\label{fig:vis_3}
\end{figure}

Markings play a crucial role in increasing driver yielding, as demonstrated in Fig.~\ref{fig:markings}. Intersections with crosswalk markings exhibit a higher percentage of driver yielding compared to driver non-yielding. Interestingly, sites with standard markings (two parallel lines) had higher yielding rates than those with continental markings, contrary to our expectations. Among the sites, 12 had no markings, 3 had standard markings, and 3 had continental markings. It is important to note the small sample size of sites with standard and continental markings.
The presence of a bike lane correlated with a decrease in driver yielding, although only 9.1\% of the interactions in the dataset occurred at sites with bike lanes, as shown in Fig.~\ref{fig:bike_lane}. As expected, sites with crosswalk signs had higher driver-yielding rates, as depicted in Fig.~\ref{fig:signage}.

\begin{figure}
    \centering 
\begin{subfigure}{0.49\textwidth}
\includegraphics[width=\linewidth]{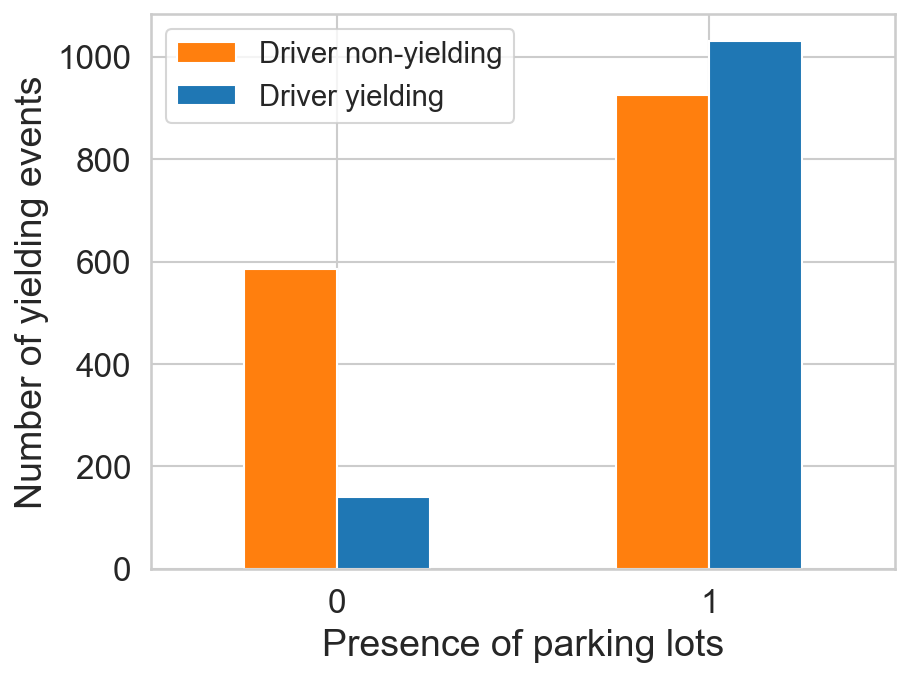}
\caption{Presence of parking lots vs. yielding. '0' denotes the absence of parking lots, while '1' indicates their presence.\label{fig:parking_lot}}
\end{subfigure}\hfil 
\begin{subfigure}{0.49\textwidth}
\includegraphics[width=\linewidth]{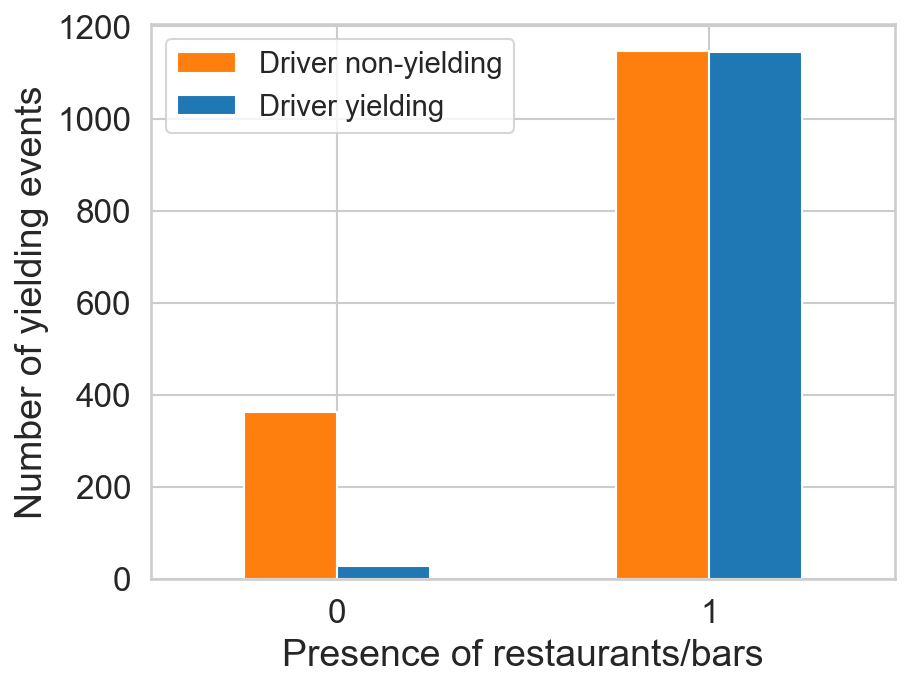}
\caption{Presence of restaurants/bars vs. yielding. '0' denotes the absence of restaurants/bars, while '1' indicates their presence.}
\end{subfigure}\hfil 
\caption{Visualization of predictor variables: Adjacent land uses}
\label{fig:vis_4}
\end{figure}

\edit{Lastly, various adjacent land use types were documented during data extraction, such as distance to a park or school and the presence of multi-family housing. Fig.~\ref{fig:vis_4} illustrates the impact of two of these land uses—restaurants/bars and parking lots within 1 block of the intersection. Sites featuring restaurants/bars and/or parking lots constituted a significant portion of the dataset, and these land uses were associated with a higher rate of driver yielding. However, in the case of parking lots, there are still more non-yielding events.}

\begin{sidewaystable}
\setlength\tabcolsep{2pt}
\begin{center}
\caption{Summary of the Data Statistics.}

\begin{tabular}{c c c c c c c c c c}

\hline
    \bf \makecell{Sites} &\bf \makecell{Total \\interactions}  & \bf \makecell{ Close\\ calls}  & \bf \makecell{Ped \\Stop midway} & \bf \makecell{Yielding \\rate (\%)}& \bf \makecell{Total \\num of ped. } & \bf \makecell{Average waiting \\time (s)}& \bf \makecell{ Average Traversing \\ time (s) } &\bf \makecell{Max waiting \\time (s)} &\bf \makecell{Min waiting\\ time (s)  } \\
    \hline
    1 & 31 & 1 & 2 & 16.13\% & 35 & 4.65 &9.58 & 24&1  \\
    2 & 100 & 0 & 13 & 43\% & 166 & 9.66&9.52 &58 &0  \\
    3 & 640 & 0 & 1 & 67.87\% & 1056 & 4.97 & 4.97 & 52 &1  \\
    4 & 143 & 0 & 0 & 5.59\% & 171 & 10.57 & 7.97&58 &0  \\
    5 & 341 & 0 & 0 & 2.93\% & 341 & 15.81 & 6.44 & 59 & 0  \\
    6 & 95 & 0 & 1 & 17.89\% & 133 & 9.98 & 9.57 &45 & 0  \\
    7 & 121 & 1 & 2 & 14.5\% & 179 & 14.15&10.25 & 58 &1  \\
    8 & 105 & 0 & 35 & 16.19\% & 144 & 10.27 &15.7&45 &1  \\
    9 & 74 & 0 & 4 & 5.41\% & 92 & 11.39&10.15 &51 &1  \\
    10 & 88 & 1 & 61 & 11.36\% & 123 & 10.17&18.01 &49 &1  \\
    11 & 20 & 0 & 1 & 0\% & 31 &13.7 &8.2 &9 &3  \\
    12 & 122 & 0 & 0 & 4.92\% & 183 & 12.58&7.78 &50 &2  \\
    13 & 98 & 1 & 8 & 59.18\% & 161 & 5.29& 7.92& 42& 1 \\
    14 & 133 & 0 & 1 & 64.47\% & 273 & 5.04& 8.18&16 &1  \\
    15 & 46 & 0 & 1 & 0\% & 72 & 18.52&8.43 &53 &2  \\
    16 & 840 & 9 & 43 & 70.36\% & 1221 &7.74 & 7.39&57 &1  \\
    17 & 127 & 3 & 9 & 17.32\% & 152 &11.84 &8.76 &54 &1  \\
    18 & 189 & 0 & 9 & 5.29\% & 252 &7.05 &14.39 &57 &1  \\
    All & 3314 & 16 & 189 & 40.38\% & 4941 &9.43 &9.23 &59 &0  \\

\hline
\end{tabular}
\label{tab:stat_locations}
\end{center}        
\end{sidewaystable}

\section{Modeling Driver Yielding}
\edit{This section outlines the statistical analysis conducted to model driver-yielding rates at 18 unsignalized intersections under study. Through the application of the developed model, along with comparative analysis against other models, we can identify significant factors influencing driver yielding at these sites and estimate the likelihood of a driver yielding under the prevailing conditions at similar locations. }

\subsection{Logistic Regression}
Logistic regression is a robust technique for binary classification that yields a model with good interpretability~\cite{kleinbaum2002logistic,ziakopoulos2020review}. It models the probability of a binary dependent variable $Y$, which in this case is the driver-yielding response that has a binary outcome - either the driver yields to the pedestrian or fails to yield. We assume a linear relationship between the independent variable $X$ and the log-odds of the event. This linear relationship is shown in~\eqref{eq:log1}, where $L$ is the log-odds, $\beta_{i}$ are the model parameters, and $P (Y = 1)$ for driver yielding and $1 - P (Y = 0)$ for driver not yielding are the probabilities of the aforementioned binary outcomes. The logistic regression model takes the form of Equation~\eqref{eq:log2} to estimate the probability of the driver yielding $P(Y=1)$, and conversely, Equation~\eqref{eq:log3} estimates the probability of the driver not yielding $1-P(Y=1)$. For the driver-pedestrian interactions considered, the outcome is either driver-yielding ($Y=1$) or non-yielding ($Y=0$). 

\begin{equation}
L(log(odds)) = ln\left ( \frac{P}{1-P} \right )=\beta_{0}+\beta_{1}x_1+\beta_{2}x_2 + \cdots +\beta_{n}x_n
\label{eq:log1}
\end{equation}

\begin{equation}
P(Y=1)=\frac{\exp(\beta_{0}+\beta_{1}x_1+\beta_{2}x_2 + \cdots +\beta_{n}x_n)}{1 + \exp(\beta_{0}+\beta_{1}x_1+\beta_{2}x_2 + \cdots +\beta_{n}x_n)}
\label{eq:log2}
\end{equation}

\begin{equation}
P(Y=0)=\frac{1}{1 + \exp(\beta_{0}+\beta_{1}x_1+\beta_{2}x_2 + \cdots +\beta_{n}x_n)}
\label{eq:log3}
\end{equation}

\begin{flushleft}
where $\beta_0$ is the constant, 
$\beta_n$ is the coefficient of the explanatory variable, 
$x_n$ is the predictor variable, and $n$ is the number of features. 
\end{flushleft}

The collected interaction data were divided into training and testing data sets, with 80\% of the data used for training the model and the remaining 20\% reserved as hold-out test data. The parameters used in the logistic model were selected through the stepwise logistic method, which is a technique used to identify the critical explanatory variables. The stepwise process combines forward and backward elimination to select the crucial variables with p-value smaller than $0.05$~\cite{bursac2008purposeful}. The selection criterion for these predictor variables relies on the Akaike Information Criterion (AIC) \cite{sakamoto1986akaike}, a measure that gauges both prediction error and model quality. Recognizing that some variables in this dataset are categorical, we implemented feature engineering to convert these into dummy variables, adopting values of either 0 or 1.

\subsection{Model for Driver-pedestrian Interaction}
Table~\ref{tab:log_results} lists the variables identified as significant and thus included in the logistic regression model of driver yielding. Based on the variable selection criterion introduced above, the vehicle speed, the yielding action of drivers in the opposite direction, the width of major crossing road, the presence of restaurants/bars/parking lots, the distance to the nearest park, and the distance to the nearest school are all significant. Note that the presence of parking refers to the large street-facing parking lots (not on-street parking) within a 1-block radius of the intersection. Therefore, the logistic regression model presented in Table~\ref{tab:log_results} is constructed using the above variables. The coefficients $\beta_n$ within the model are determined using maximum-likelihood estimation, an iterative procedure aimed at minimizing the error in the predicted probabilities. These coefficients $\beta_n$ articulate the magnitude of the influence each predictor variable has on the binary outcome of driver yielding or not yielding. Additionally, the standard error provides an estimate of each coefficient's variability, while the Z-score conveys the deviation from the mean in units of standard deviation. A test statistic, the p-value, reflects the probability of obtaining extreme results from the statistical analysis. Specifically, a low p-value signifies a predictor's likely relevance to the model.

\begin{table}
\begin{center}
\captionof{table}{Logistic regression model results with selected variables.}
\begin{tabular}{cccccc}
\hline
\textbf{Variables} & \textbf{coefficient} & \textbf{std err} & \textbf{z} & \textbf{P$> |$z$|$} & \textbf{Effect}   \\
\hline
\textbf{Crossing width (major)}   &   0.098  &   0.01     &  9.383 &         0.000     & Positive\\
\textbf{Presence of Restaurants/Bars}    & 1.8315  &0.409 & 4.478  &         0.000    & Positive\\

\bf{Vehicle speed} & -0.2402  &        0.011     & -20.989 &     0.000  &  Negative     \\
\textbf{Dist. to nearest school} & -0.4459 &   0.163   &  -2.736  &         0.006    & Negative   \\
\textbf{Presence of parking lots}    &  -1.6236  &  0.263  &-6.183  &         0.000   & Negative   \\
\textbf{Opposite direction yield}  &-1.6438  &   0.225  &  -7.294  & 0.000  & Negative  \\

\textbf{Dist. to nearest park}   &  -4.0946  &  0.996    &  -4.111  &         0.000  & Negative \\
\hline
\label{tab:log_results}
\end{tabular}
\end{center}
\end{table}

\begin{figure}
    \centering 
\begin{subfigure}{0.49\textwidth}
\includegraphics[width=\linewidth]{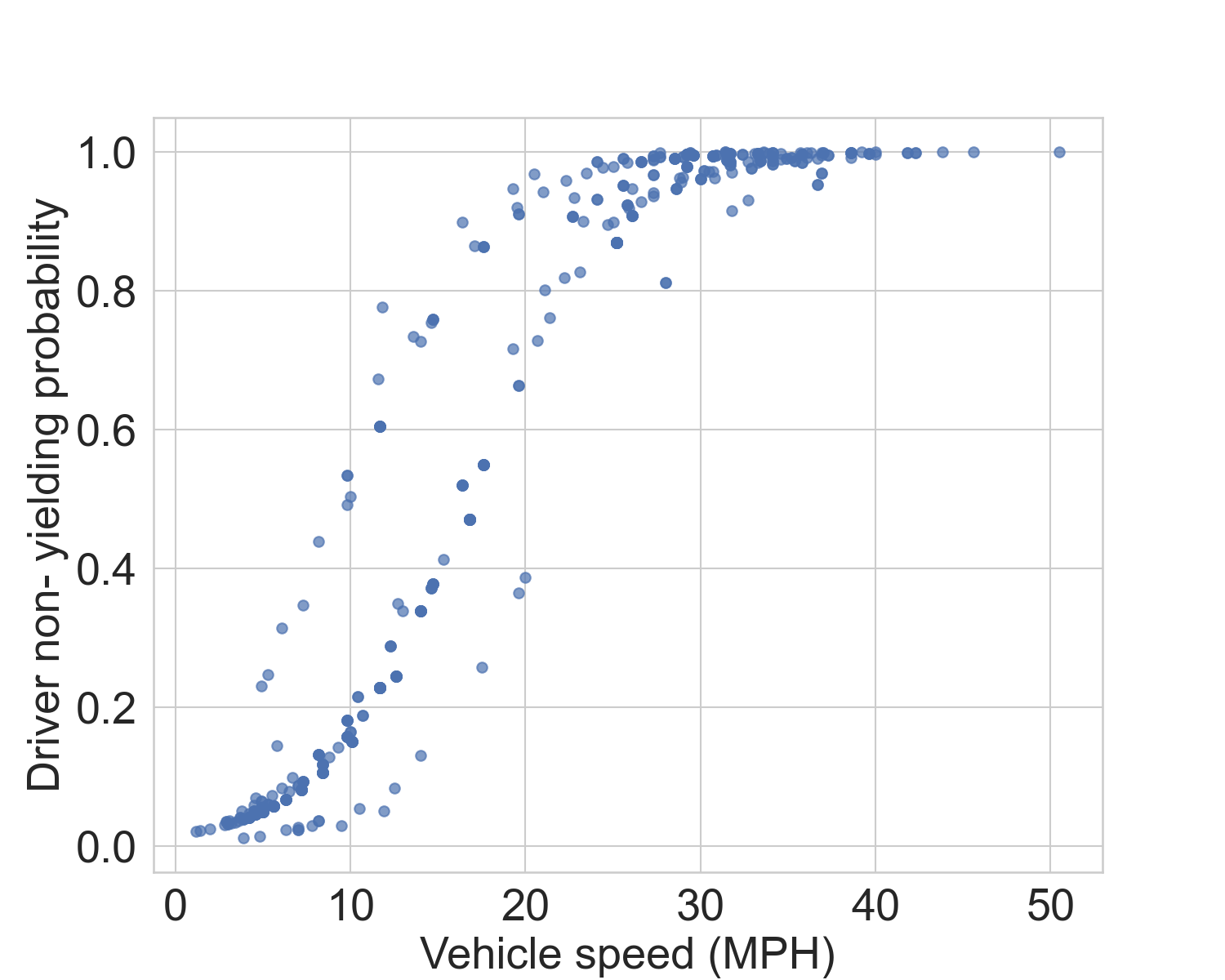}
\caption{Vehicle driving speed vs. Driver non-yielding probability. \label{fig:prob_speed}}
\end{subfigure}\hfil 

\caption{Predictors vs. yielding probability.}
\label{fig:model_prob}
\end{figure}

Based on the result in Table \ref{tab:log_results}, all the selected variables are significantly crucial at the 95\% confidence level. The odds ratios for all the significant variables are also calculated. The odds ratio represents the change in the probability of an explanatory variable influencing the difference in the response variable. The odds of driver yielding is $exp(-0.2354) =  0.79$ times lower with each unit (MPH) increase in vehicle speed. Specifically, each unit increase in vehicle speed is associated with a 21\% $(1-0.79 = 0.21)$ reduction in the relative probability of driver yielding, assuming other variables remain fixed (refer to equation~\eqref{eq:log1}). Vehicle speed emerged as the most crucial variable based on the Z-score. Therefore, we develop Fig.~\ref{fig:model_prob} to visually depict how the model predicts the probability of driver non-yielding based on one of the independent variables: vehicle speed. 

In addition, the study revealed that the yielding action of drivers in the opposite direction adversely affects driver yielding, reducing the relative probability by 80.7\%. Conversely, wider major road crossings positively influence driver yielding. Each additional foot in road width correlates with a 10.3\% increase in the likelihood of drivers yielding.

In the case of built environment-related factors, this result exhibits that the odds ratio for the presence of restaurants/bars is 6.24 relative to not present, and 0.197 for the presence of a parking lot than not. Thus, a site with restaurants/bars has 544\% $(6.24-1 = 5.24)$ greater relative driver-yielding probability than a site that does not have restaurants/bars. Additionally, the proximity of parks or schools appears to have a positive effect on driver-yielding behavior. Specifically, for each mile closer to the nearest park or school, the probability of driver yielding increases by 98\% and 36\%, respectively. This result indicates the significance of adjacent built environment context on the driver-yielding behavior.

\edit{In our analysis, certain factors such as pedestrian group size and weather were found to be insignificant. It is also worth noting that large standard errors for some independent variables might be a consequence of imbalanced data in this study. Specifically, data were collected from 18 locations, but the total number of events varies among these locations, as illustrated in Table~\ref{tab:stat_locations}. This variation may have contributed to the observed disparities and should be considered in interpreting the results.}

\subsection{Model Evaluation}
The logistic model's confusion matrix is shown in Table~\ref{tab:confusion_matrix}. It's evaluated using three metrics: accuracy, precision, and recall, defined in equations \eqref{eq:precision} and \eqref{eq:recall}. Precision quantifies the ratio of correctly predicted positive instances to the total number of positive predictions made by the model. Recall, on the other hand, is the proportion of correctly predicted positive instances to the total actual positive instances. Notably, the models tend to have more false positives (Type \rom{1} Error) than false negatives (Type \rom{2} Error), resulting in higher recall values than precision. This trend indicates that the model is more prone to falsely predict driver yielding when the driver is not actually yielding.

\begin{equation}
Precision = \frac{\text{True Positive}}{\text{True Positive+False Positive}}
\label{eq:precision}
\end{equation}

\begin{equation}
Recall = \frac{\text{True Positive}}{\text{True Positive+False Negative}}
\label{eq:recall}
\end{equation}

\begin{equation}
Accuracy = \frac{\text{(True Positive+True Negative)}}{\text{(True Positive+True Negative+False Positive+False Negative)}} 
\label{eq:accuracy}
\end{equation}

\begin{figure}
     \centering
        \includegraphics[scale = 0.5]{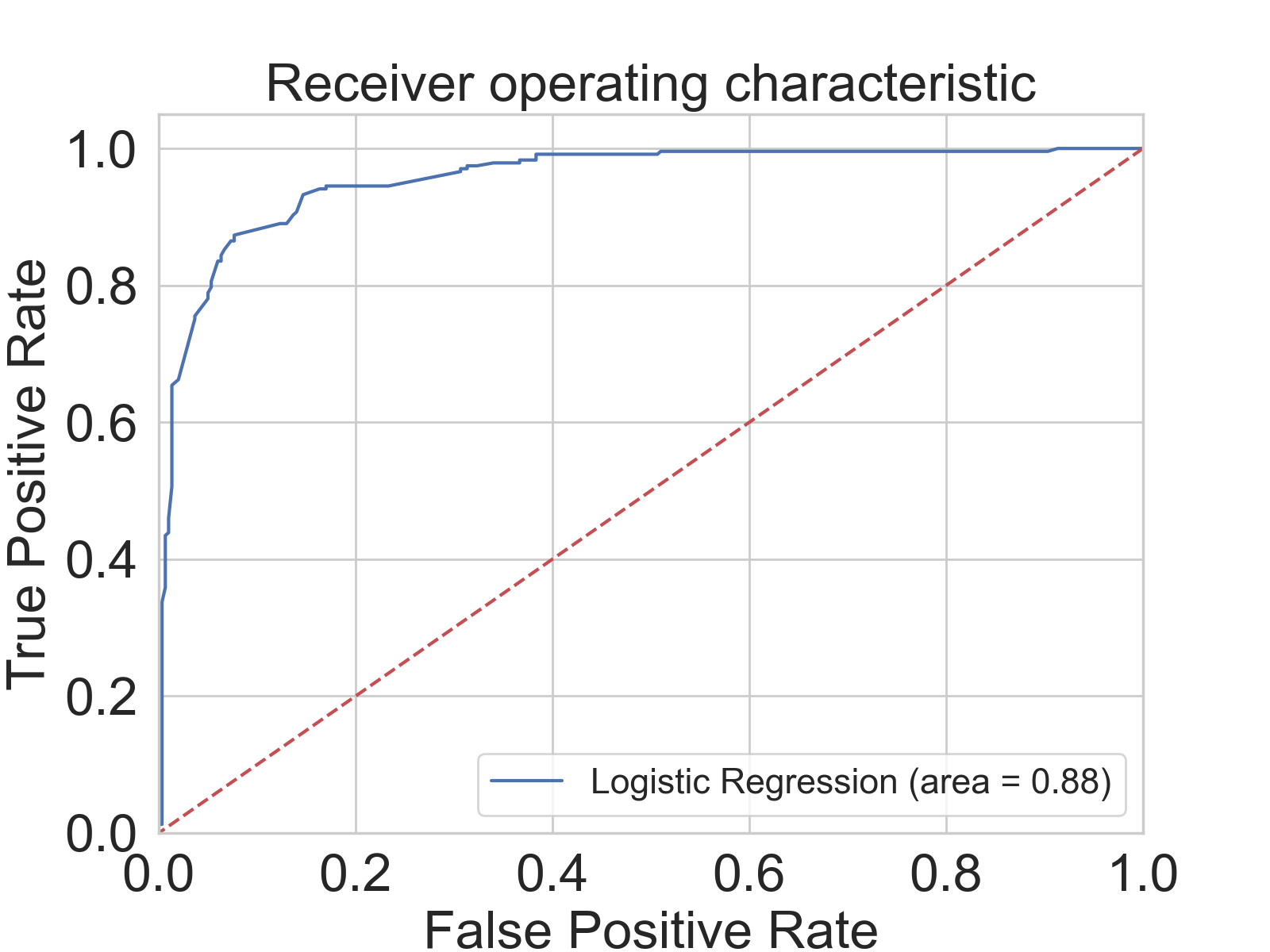}
     \caption{ROC curves for the logistic regression model.}
     \label{fig:ROCs}
\end{figure}

Fig.~\ref{fig:ROCs} displays the Receiver Operating Characteristic (ROC) curves, which are plotted with respect to true positive and false positive rates. The ROC curve is a performance measurement for binary classification at various threshold settings. For our model, the area under the ROC curve is 0.88 for testing data, closely following the left-hand border and top border of the ROC space. This signifies that the model exhibits good performance, adeptly balancing sensitivity and specificity, and also underscores the importance of the selected variables in predicting the outcome of driver-pedestrian interactions.

\begin{table}
\begin{center}
    \captionof{table}{Model Performance in Test Data.}
\begin{tabular}{cccccc}
\hline
\multirow{2}{*}{\textbf{Models}} & \multirow{2}{*}{\textbf{Actual}} &  \multicolumn{2}{c}{\textbf{Predicted}} \\
& & \textbf{Driver not yield} & \textbf{Driver yield} & \  \\
\hline

\multirow{5}{*}{\textbf{\makecell{Model with \\selected features}}} & \textbf{Driver not yield} &       258  &       42      \\
& \textbf{Driver yield}  &     22 &        211       \\
& \textbf{Accuracy Score}  &     \multicolumn{2}{c}{0.88} \\
& \textbf{Precision}  &     \multicolumn{2}{c}{0.84} \\
& \textbf{Recall}  &     \multicolumn{2}{c}{0.91} \\
\hline

\label{tab:confusion_matrix}
\end{tabular}
\end{center}
\end{table}

Lastly, we compare the accuracy in the test data of our logistic regression model with other machine learning models, including \textit{Support Vector Machine} (0.88), \textit{Random Forest} (0.885), and \textit{Neural Network} (0.89). Though the performances are similar, the logistic regression model stands out for its superior interpretability. Therefore, we choose logistic regression as our primary model.

\section{Discussion and Conclusion}
In summary, many factors influence whether or not a driver will yield to a pedestrian at an unsignalized intersection. This research endeavors to gather field data from various intersections and carry out in-depth analyses to pinpoint the most crucial factors, including built environment factors, governing a driver's yielding behavior. Consequently, we employ the custom-built TIM sensor platform, developed by the Minnesota Traffic Observatory, to obtain vehicle-pedestrian video data. We then construct logistic regression models for modeling driver yielding behavior and utilize these models to evaluate prediction accuracy and identify the most influential variables in determining a driver's decision to yield to pedestrians. The primary findings of our study can be summarized as follows:

\begin{itemize}

\item The collection and publication of driver-pedestrian yielding data from 18 intersections across Minnesota, constituting one of the most extensive datasets of this type publicly available in the U.S. 

\item The construction of the predictive model leveraging selected variables for optimal explanation and potential improvement. These models help in evaluating the variables that most significantly influence driver-yielding behavior.

\item Surrounding built environment factors have an influence on driver-yielding behavior, which represents a significant new knowledge contribution to the literature on driver-yielding behavior.
\end{itemize}

Table~\ref{tab:log_results} delineates the specific variables considered in predicting driver yielding. The findings of this study highlight that vehicle speed is the most influential predictor of driver-yielding behavior. The faster the vehicle's speed at a point near the intersection, the less likely the driver is to yield. This result is intuitive, as high-speed vehicles find it more challenging to stop or slow down for pedestrians compared to lower-speed vehicles, resulting in comparatively lower yielding rates. This result is also supported by previous studies \cite{zheng2015modeling,schroeder2011event,salamati2013event,bai2013comparative,silvano2016analysis}, and emphasizes the significant role of vehicle speed in forecasting driver behavior.

\editnew{The model's findings indicate that a wider major road crossing width is associated with higher driver-yielding behavior, which differs from the previous study~\cite{zhao2020modeling}. On the other hand, the study found that when drivers in the opposite direction yield, it reduces the probability of driver yielding, a result contrary to the findings of certain previous studies~\cite{schroeder2011event,akin2010neural}. One possible explanation for this counterintuitive result is that drivers may assume the pedestrian has already been accommodated by the opposite vehicle and is no longer in their path. This assumption could lead them to believe that further yielding is unnecessary, prompting them to proceed without stopping. }

\editnew{We have categorized the variables into two groups: those related to specific events (e.g., vehicle speed, pedestrian type, interaction type), which are generally beyond immediate control, and those related to the intersection’s built environment. While built environment factors may not be easily modified, our study provides valuable insights that can guide local agencies in making informed decisions for future intersection designs or modifications. It is important to note that at the beginning of our study, we included land use as a feature in our model. However, due to the limited variability in land use characteristics among the 18 intersections we studied, this variable did not prove to be significant.}

Built environment factors significantly influence driver-yielding behavior. The presence of restaurants/bars, closeness to schools, and proximity to parks increase driver-yielding rates. These locations typically experience higher levels of pedestrian activity. Therefore, various crash reduction countermeasures, such as speed control, provision of sidewalks, signage, and pedestrian-friendly infrastructure, might already be available in these highly active areas to enhance pedestrian safety, potentially increasing the probability of drivers yielding. However, the presence of parking lots reduces the probability of driver yielding. This is likely due to parking lots reducing the visibility of pedestrians, making it difficult for drivers to anticipate pedestrian arrivals for crossing. Consequently, drivers may not be mentally prepared to slow down or stop for pedestrians, thereby decreasing driver yielding rates.

\editnew{In case of limitations and further study scope, the dataset covers 18 intersections in Minnesota, its applicability to other regions might vary. Other limitations include data imbalances among locations, a limited scope of studied variables and conditions, and the absence of certain pedestrian features, such as gender and race, which could affect the generalizability of the results. Future research should consider geographical expansion, the incorporation of additional relevant variables, and the use of diverse machine learning models. Data fusion techniques that incorporate multi-source data could be particularly useful in developing models with improved interpretability. Additionally, analyzing crash frequency and severity connecting with the yielding rate, could be valuable next steps.}


\section{Practical Applications}

Based on the study's findings and discussion, several actionable recommendations can be made to positively influence driver-yielding behavior. Implementation of posted speed limits, installation of speed control measures, and enforcement of speed limits are necessary to control vehicle speed at unsignalized intersections. Replicating effective features found near high pedestrian activity areas (school/park/bar/restaurants), such as pedestrian crossing markings, high-quality pedestrian infrastructure, traffic calming measures, and improved signage and marking, in other pedestrian crash-prone areas, including those near parking lots, can help increase driver yielding behavior. Furthermore, installing pedestrian signals (push buttons) at unsignalized intersections, especially those near parking lots, can increase driver-yielding rates. Furthermore, incorporating comprehensive pedestrian safety training into driver education programs and providing ongoing training for licensed drivers to reinforce the importance of yielding to pedestrians would further enhance driver yielding rates.

These recommendations align with Vision Zero's goals, fostering a safe and forgiving road environment. The implications of this study extend beyond Minnesota, potentially benefiting traffic safety initiatives throughout the United States. Furthermore, the findings are likely to be valuable in the design and development of automated vehicles in the future.

\section*{Author Contributions}

The authors confirm their contribution to the paper as follows: study conception and design: T. Li and R. Stern; data collection: T. Li, J. Klavins; analysis and interpretation of results: T. Li, J. Klavins, T. Xu, N. Zafri, and R. Stern; draft manuscript preparation: T. Li, J. Klavins, T. Xu, N. Zafri, and R. Stern. All authors reviewed the results and approved the final version of the manuscript. The authors do not have any conflicts of interest to declare.

\section*{Acknowledgment}
Full results and details of the model and the data are available on request. The authors would like to thank Christopher Cheong and Shirley Shiqin Liu for their constructive criticism of the manuscript. 

\section*{Declaration of Conflicting Interests}
The author(s) declared no potential conflicts of interest with respect to the research, authorship, and/or publication of this article.

\section*{Funding}
This work is supported by the Minnesota Department of Transportation under contract No.1036210. T. Li acknowledges the support of the Dwight David Eisenhower Graduate Fellowship from the Federal Highway Administration. 

\bibliographystyle{unsrt}  
\bibliography{refs}  

\section{Appendices}

\noindent\textbf{Appendix A. Site characteristics}\medskip

This section explains some of the important site characteristics at each of the 18 locations that were studied. Each site had a mounted camera facing the crosswalk(s) that were analyzed. Below are the detailed descriptions of the 18 locations with their relevant attributes after the equipment was in place. 
Table~\ref{tab:Variables_description} shows the most commonly collected variables in the study, but there are also more variables in our published dataset. 

\begin{enumerate}
\item Site 1 is a T-intersection with four lanes at the main street and a median in the middle of the main street. The camera was mounted onto a light post on the northwest side of the intersection. The speed limit at this site is 35 mph. This site was close to many small businesses and food establishments. This site had a bus stop on only one side of the major road. It is also located in front of a roundabout and features four lanes that pedestrians must cross to get to the other side of the road. This intersection also has a median with ADA pads installed within it. This site experiences an AADT of about 14600 (2016) on the major road.

\item Site 2 is a T-intersection with two lanes at the main street and a median in the middle of the road with a speed limit of 35mph. The camera was mounted onto a light post that was pretty close to the intersection. Its wide median made the road width seem relatively narrow. This intersection had a painted marked crosswalk and had ADA pads on each side of the street and in the median. The second crosswalk was unmarked and had one ADA pad. Another thing that made this site unique is that it contained an identical walkway one street down with all the same characteristics except for a marked crosswalk. This site had a major road AADT of 14800 (2016).

\item Site 3 is a T-intersection with two lanes at the main street and a median in the middle of the major road. The speed limit at this site was 35 mph. The camera was mounted onto a light post a reasonable distance away from the intersection. This site featured two crosswalks, the first identical to the previous site, besides not being painted. The second crosswalk was unmarked and located a bit closer to the camera. This site also had two bus stops on each side of the major road. The AADT associated with the major road at this site was also 14800 (2016).
    
\item Site 4 is a four-way intersection with three lanes on the main street, one of which is a turn lane for both directions of the major road. The camera was mounted onto a light post just north of the two crosswalks. There was also a bus stop near the end of the crosswalk on each side of the major road. This site was located in a residential area surrounded by many houses. This intersection had two marked continental crosswalks on both sides of the minor road. It also had a turn lane, a short median on each side of the minor road, and a posted speed limit of 35mph. The AADT of the major street was roughly 10700 (2016).
    
\item Site 5 is a four-way intersection with four lanes at the main street, two of which were turn lanes for each direction on the major road. The speed limit on the major road is 35mph. The camera was mounted onto a road sign southwest of the observed crosswalk. This site was located near many businesses and restaurants on the north side of the street and houses on the south side of the major road. This site had an unmarked crosswalk, and the camera that was set up here could only capture this section of the intersection because there were not many ideal spots to mount the camera. The AADT of the major road was approximately 10500 (2018).
    
\item  Site 6 is a four-way intersection with two lanes at the main street and parking on each side of the major road. The camera was mounted onto a light post at the intersection's southwest corner. This site was located in a busy area of the city next to shops and attractions. These characteristics generated a very high volume of both drivers and pedestrians. This site has one bus stop in the southeast corner of the intersection and a posted speed limit of 30mph. The intersection featured four marked, standard crosswalks for pedestrians to cross in any direction. The two crosswalks over the major street were observed in this study. The AADT of the major street was 6200 (2016). 
    
\item Site 7 is a T-intersection with two lanes at the main street and parking on either side of the major road. The camera was mounted onto a light post at the intersection's southwest corner. This site was also located in a busy area of the city next to many restaurants and shops and had a speed limit of 30mph. Given these characteristics, this intersection also had a very high pedestrian volume despite only having one marked, standard crosswalk on the northern side of the intersection. The AADT of the major road at this site was 3300 (2012).
    
\item Site 8 is a four-way intersection with two lanes on the main street and a center turning lane for left-turning traffic on either side of the road. The camera was mounted onto a sign due North of the intersection. This site was next to many residential, single-family homes and had a posted speed limit of 35mph. In addition, this site featured two bus stops near the intersection and had no marked crosswalks. This site also had a  large volume of vehicles throughout the day, leading to a major street AADT of about 10200 (2018). 
    
\item Site 9 is a four-way intersection with two lanes at the main street. The camera was mounted onto a light post to the northwest of the intersection. This intersection had a speed limit of 30mph and a single bus stop at its southwest corner. This site was next to a new apartment complex and a few shops and restaurants, including a convenience store. This intersection had no marked crosswalks on the street, which caused pedestrians to wait a while before crossing. This site had a major road AADT of about 5400 (2020). 
    
\item Site 10 is a four-way intersection with four lanes at the main street. The camera was mounted onto a light post a good distance to the northwest of the intersection. Single and multifamily houses and buildings surrounded three-quarters of this intersection. The other quarter contained both restaurants and shops. This site featured two marked continental crosswalks over the major road. This intersection also had two bus stops, one on each side of the major road, and a speed limit of 35mph. The width of this intersection and the number of lanes generally made it a challenge for pedestrians trying to cross. This site had a very high vehicle volume resulting in an AADT of about 18800 (2017). 
    
\item Site 11 is a four-way intersection with four lanes at the main street. The camera was mounted onto a light post to the northeast of the intersection. This intersection had some multifamily residential developments under construction at the time and single-family homes on the opposite side of the major road. The major road had a posted speed limit of 30mph. This site did not have marked crosswalks and had a very high vehicle volume. The AADT at this site was 18100 (2018). 
    
\item Site 12 is a four-way intersection with two lanes at the main street. The camera was mounted onto a light post near the northeast corner of the intersection, which only allowed for the far crosswalk on the western side of the intersection to be studied. This intersection was near single-family homes and a few shops and restaurants. The speed limit at this intersection is 30mph. There were no marked crosswalks at this intersection, and this site experienced a pretty low pedestrian volume. The AADT associated with the major road was 10300 (2017). 
     
\item Site 13 is a four-way intersection with two lanes at the main street. The camera was mounted onto a tree southeast of the intersection. This site only had one marked crosswalk crossing the minor street but no marked crosswalks across the major road. This site was surrounded by single-family residential homes and one larger multifamily apartment complex. This intersection had two bus stops on each side of the major road and had a posted speed limit of 30mph. This intersection experienced a decent amount of pedestrian volume and had an AADT of 7400 (2019).

\item Site 14 is a four-way intersection with two lanes at the main street and a major road speed limit of 30mph. The camera was mounted onto a light post southeast of the intersection. This intersection was near both single and multifamily homes and buildings, including a park on the west side of the intersection. There were no marked crosswalks but a high volume of pedestrians trying to cross the major road. The AADT associated with this site was 7100(2017). 
    
\item Site 15 is a four-way intersection with two lanes at the main street. The camera was mounted onto a light post on the northeast corner of the intersection. This site was near single-family houses and commercial buildings and had a speed limit of 30mph on the major road. This intersection had no marked crosswalks and a high vehicle volume. This site had two bus stops on each side of the major road and had an AADT of about 16400 (2017). 
    
\item Site 16 is located on a two-lane roadway with a center turning lane for left-turning traffic. The posted speed for this intersection is 30mph. The adjacent land use in the area around the corner is primarily commercial, with many shops and restaurants that have on-street parking on the primary road. These features amplified the pedestrian volume near and at this intersection. In addition, the intersection is a two-way stop-controlled on the minor road. The AADT on the major road at this site was 12100 (2018). 

\item Site 17 is located on a four-lane suburban road with a posted speed limit of 30 mph. This intersection does not have on-street parking. The adjacent land use consists of strip malls and several commercial buildings on the main road on both sides of the intersection. This site had two bus stops on each side of the major road. This intersection had a major road AADT of about 12800 (2016) and a minor road AADT of about 1350 (2013). 

\item Site 18 is located on a large, state-run 4-lane road that is two-way stop-controlled on the minor roadway. The posted speed limit at this intersection for the main road (subject road being studied) is 30 mph. The built-in environment is a mixed residential and industrial area with adjacent commercial zoning. This site featured a major road AADT of about 12500 (2017).
\end{enumerate}

\begin{figure}
    \centering
    \begin{subfigure}{0.3\textwidth}
        \centering
        \includegraphics[width=\linewidth]{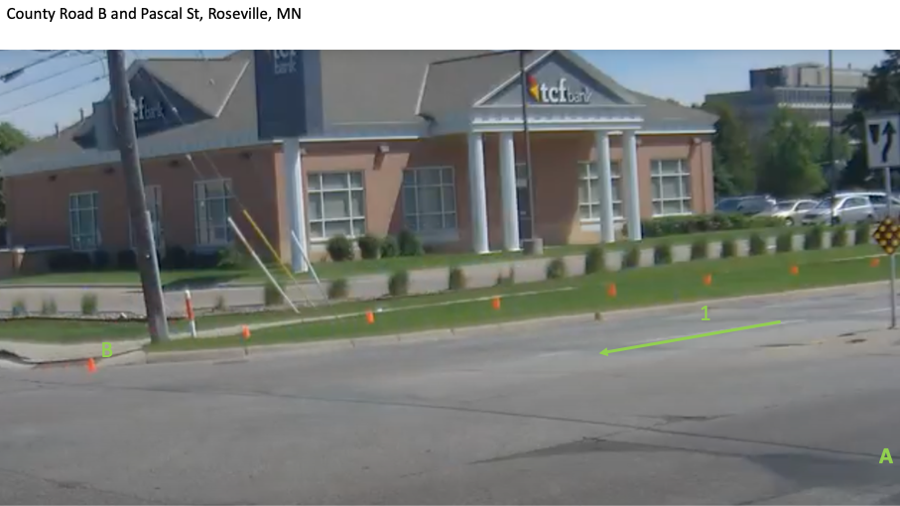}
        \caption{Site 1}
    \end{subfigure}%
    \hfill
    \begin{subfigure}{0.3\textwidth}
        \centering
        \includegraphics[width=\linewidth]{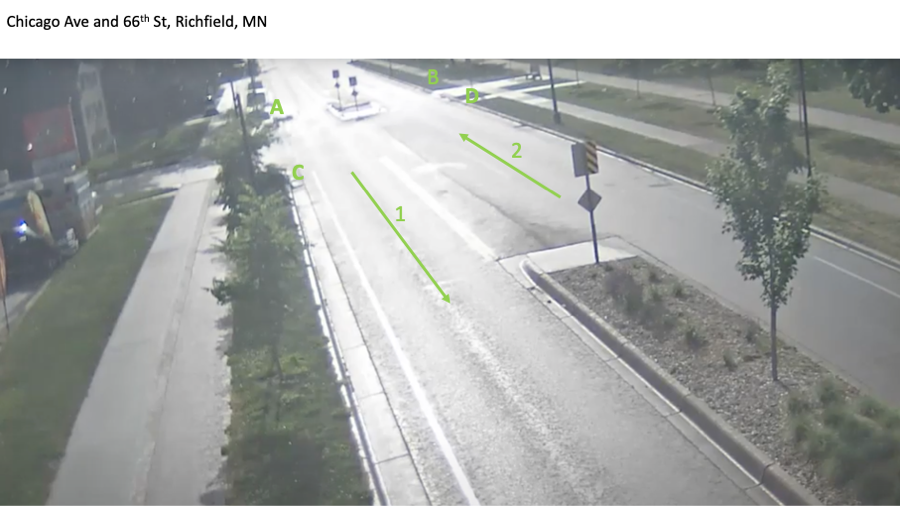}
        \caption{Site 2}
    \end{subfigure}%
    \hfill
    \begin{subfigure}{0.3\textwidth}
        \centering
        \includegraphics[width=\linewidth]{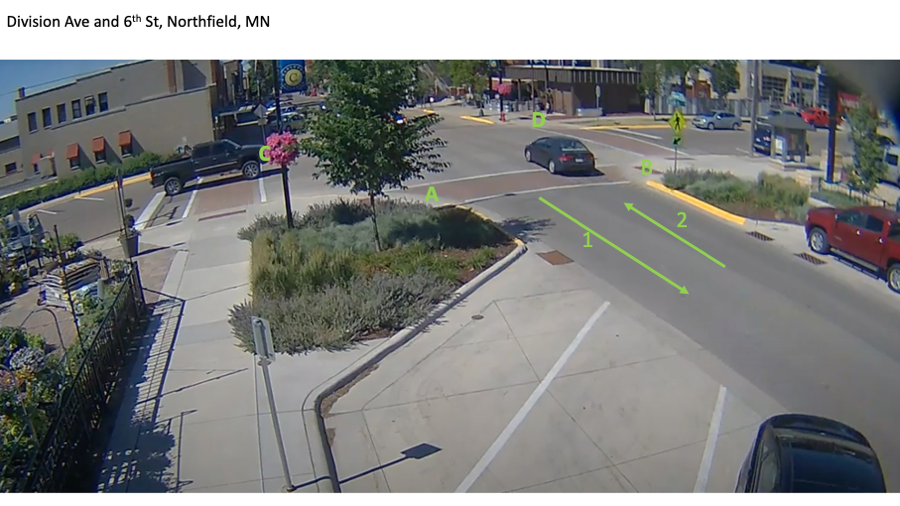}
        \caption{Site 3}
    \end{subfigure}

    \vspace{0.5cm}

    \begin{subfigure}{0.3\textwidth}
        \centering
        \includegraphics[width=\linewidth]{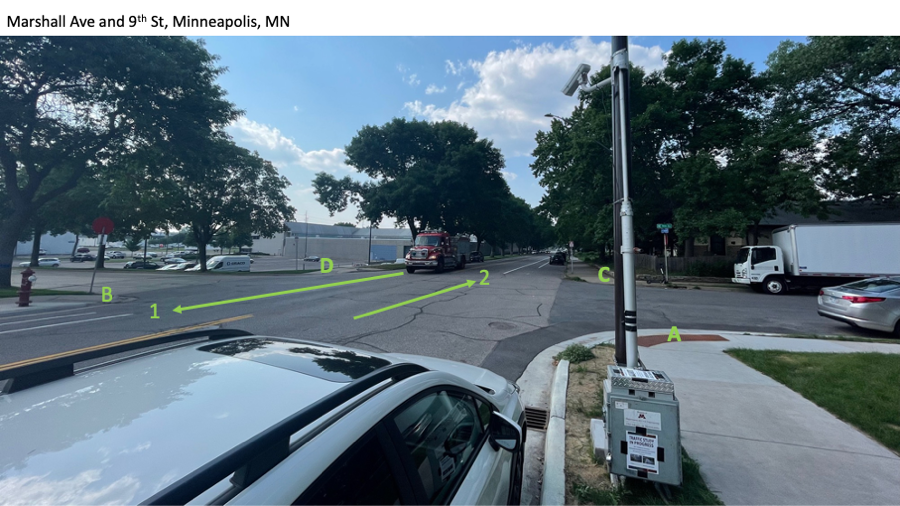}
        \caption{Site 4}
    \end{subfigure}%
    \hfill
    \begin{subfigure}{0.3\textwidth}
        \centering
        \includegraphics[width=\linewidth]{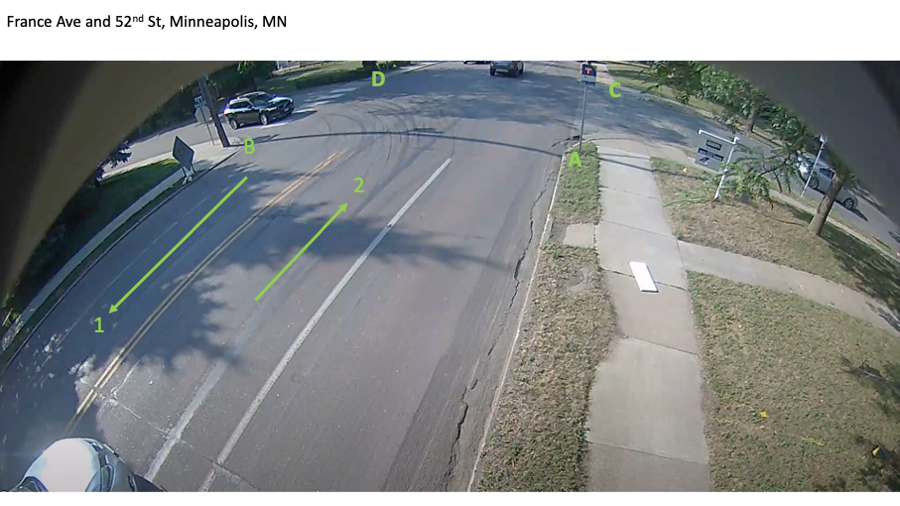}
        \caption{Site 5}
    \end{subfigure}%
    \hfill
    \begin{subfigure}{0.3\textwidth}
        \centering
        \includegraphics[width=\linewidth]{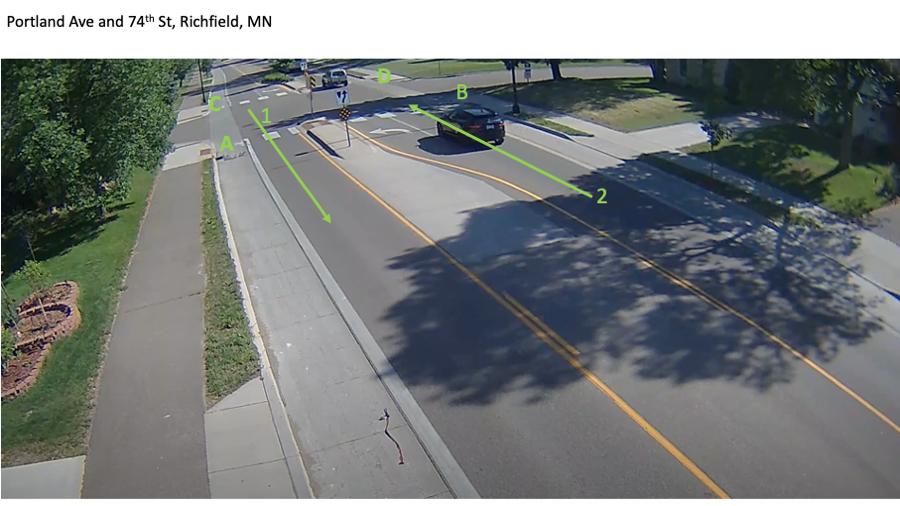}
        \caption{Site 6}
    \end{subfigure}

    \vspace{0.5cm}

    \begin{subfigure}{0.3\textwidth}
        \centering
        \includegraphics[width=\linewidth]{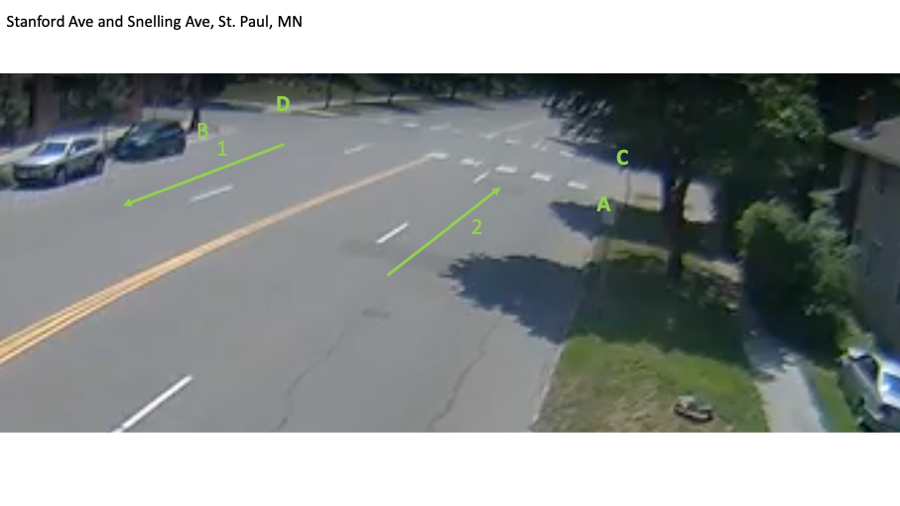}
        \caption{Site 7}
    \end{subfigure}%
    \hfill
    \begin{subfigure}{0.3\textwidth}
        \centering
        \includegraphics[width=\linewidth]{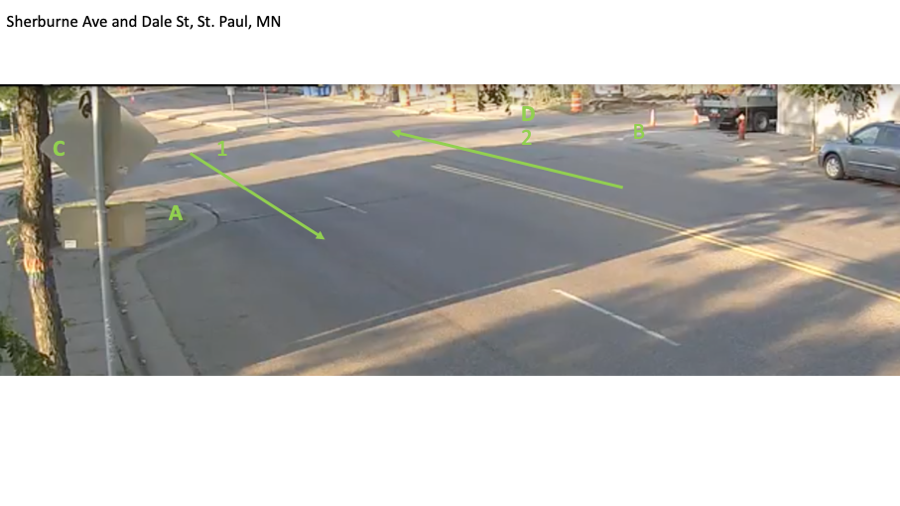}
        \caption{Site 8}
    \end{subfigure}%
    \hfill
    \begin{subfigure}{0.3\textwidth}
        \centering
        \includegraphics[width=\linewidth]{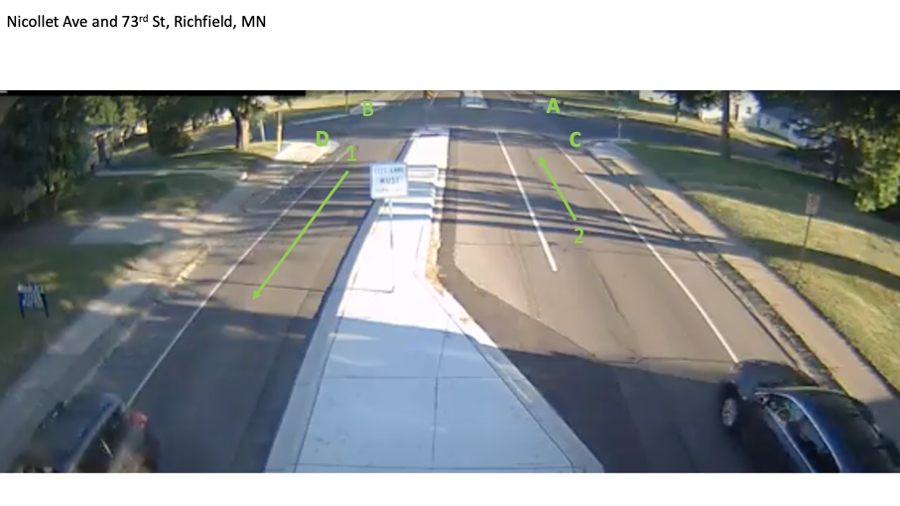}
        \caption{Site 9}
    \end{subfigure}

    \caption{Views of the study site: panel 1}
    \label{fig:array}
\end{figure}

\begin{figure}
    \centering
    \begin{subfigure}{0.3\textwidth}
        \centering
        \includegraphics[width=\linewidth]{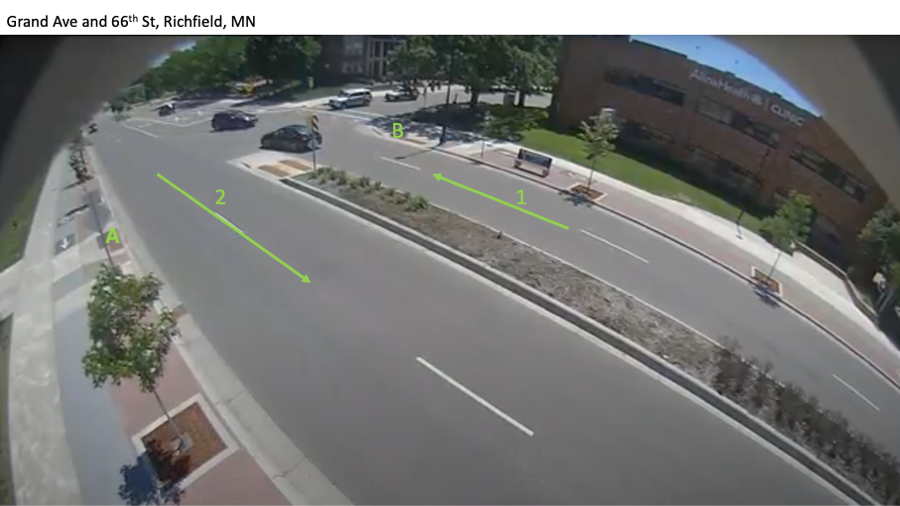}
        \caption{Site 10}
    \end{subfigure}%
    \hfill
    \begin{subfigure}{0.3\textwidth}
        \centering
        \includegraphics[width=\linewidth]{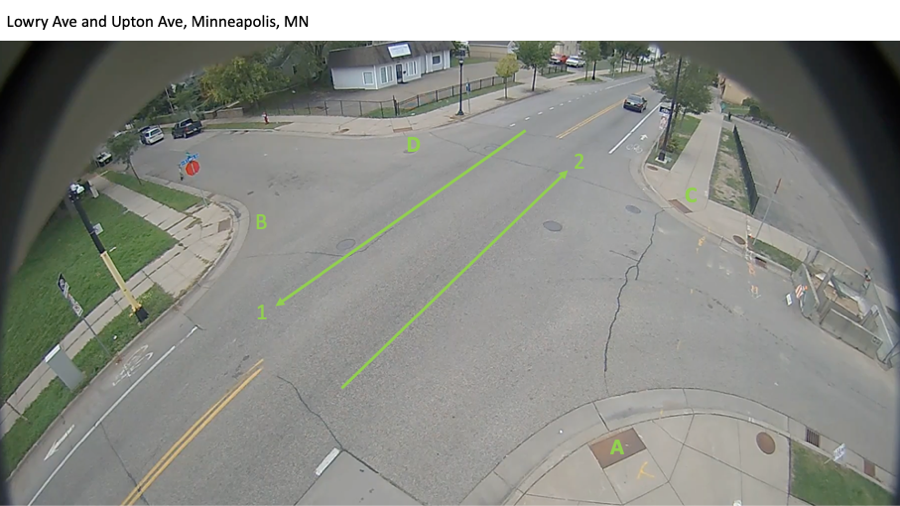}
        \caption{Site 11}
    \end{subfigure}%
    \hfill
    \begin{subfigure}{0.3\textwidth}
        \centering
        \includegraphics[width=\linewidth]{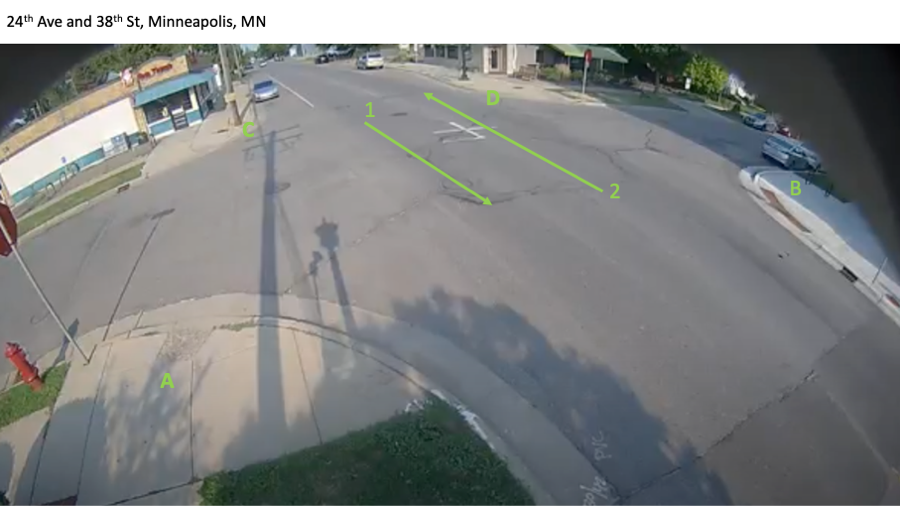}
        \caption{Site 12}
    \end{subfigure}

    \vspace{0.5cm}

    \begin{subfigure}{0.3\textwidth}
        \centering
        \includegraphics[width=\linewidth]{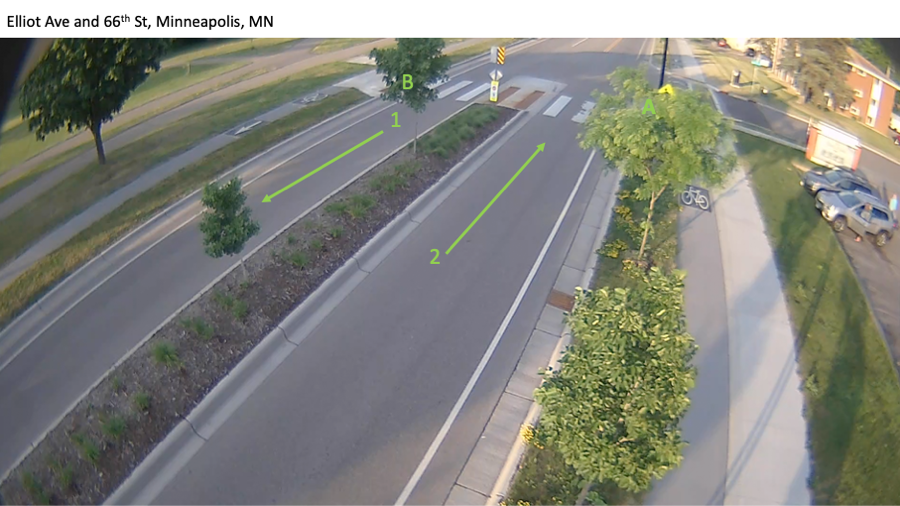}
        \caption{Site 13}
    \end{subfigure}%
    \hfill
    \begin{subfigure}{0.3\textwidth}
        \centering
        \includegraphics[width=\linewidth]{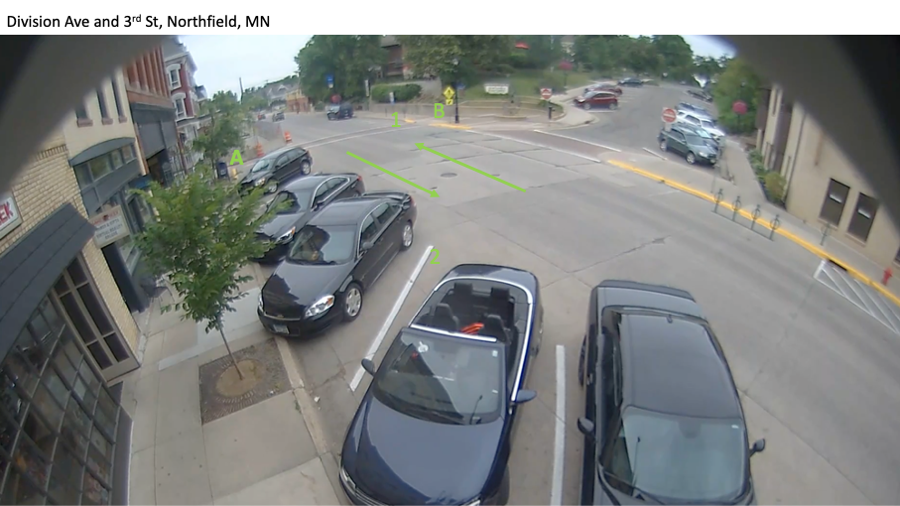}
        \caption{Site 14}
    \end{subfigure}%
    \hfill
    \begin{subfigure}{0.3\textwidth}
        \centering
        \includegraphics[width=\linewidth]{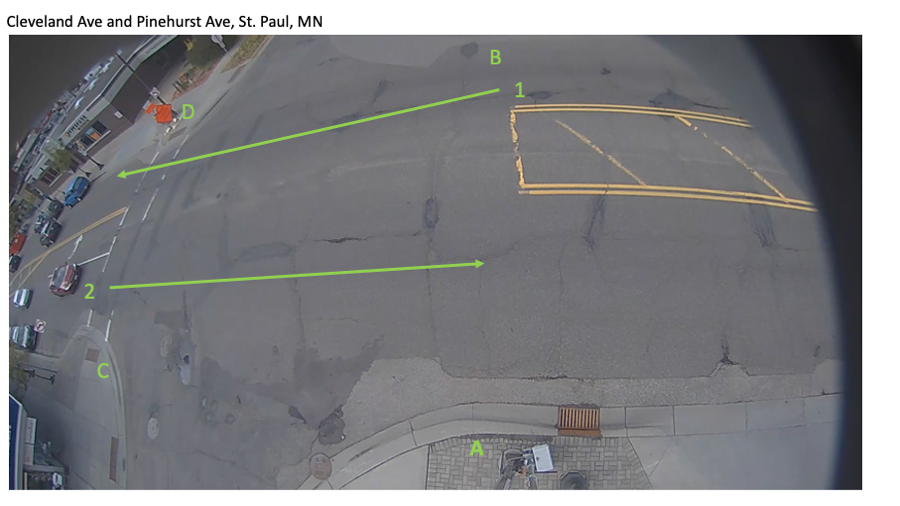}
        \caption{Site 15}
    \end{subfigure}

    \vspace{0.5cm}

    \begin{subfigure}{0.3\textwidth}
        \centering
        \includegraphics[width=\linewidth]{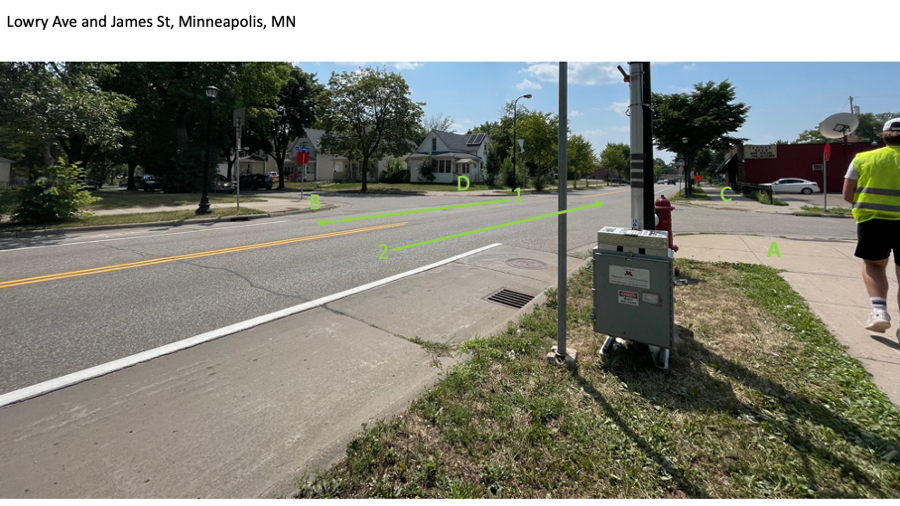}
        \caption{Site 16}
    \end{subfigure}%
    \hfill
    \begin{subfigure}{0.3\textwidth}
        \centering
        \includegraphics[width=\linewidth]{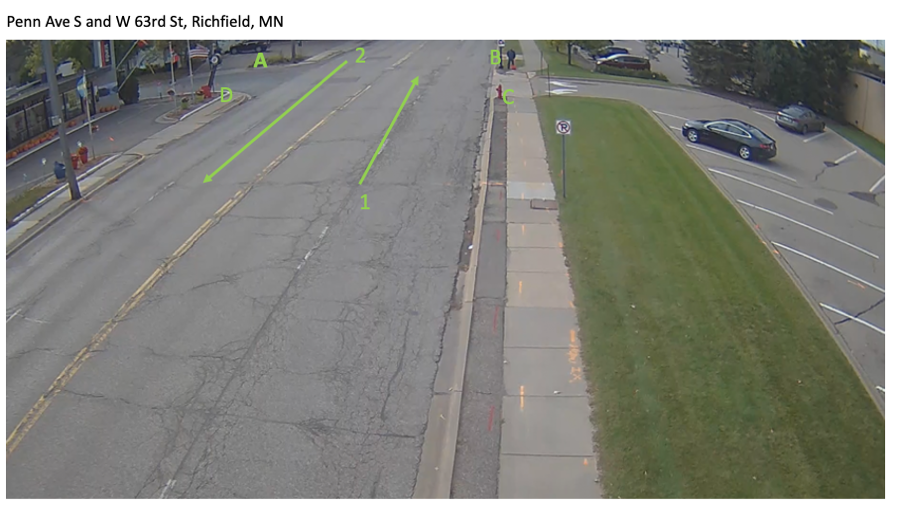}
        \caption{Site 17}
    \end{subfigure}%
    \hfill
    \begin{subfigure}{0.3\textwidth}
        \centering
        \includegraphics[width=\linewidth]{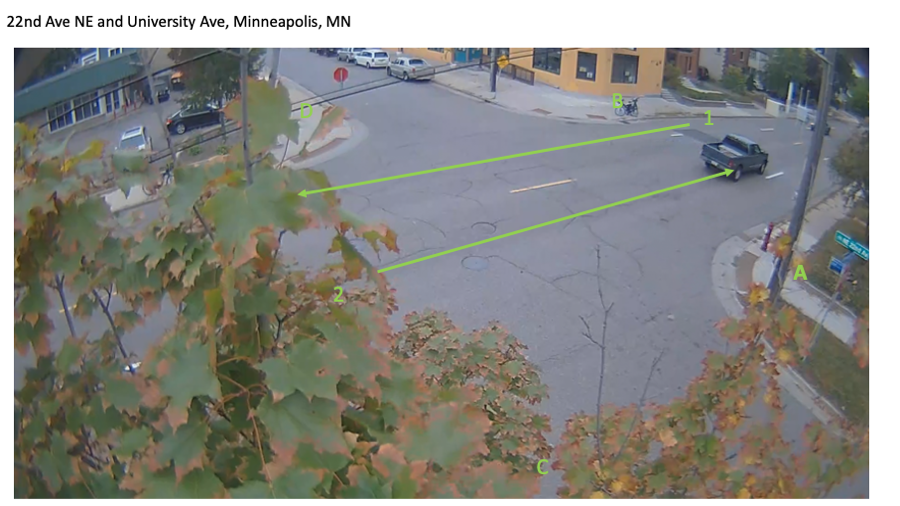}
        \caption{Site 18}
    \end{subfigure}

    \caption{Views of the study site: panel 2}
    \label{fig:array2}
\end{figure}

\begin{table}[ht]
\setlength\tabcolsep{2pt}
  \centering
  \caption{Variables collected in this study.}
  \scalebox{0.87}{
    \begin{tabular}{cccc}
    \toprule
    \bf Response Variable & \bf \makecell{Total Number\\(Cleaned data)} & \bf Ratio (P/V) \\
    \midrule
    \multirow{2}{*}{\makecell{Yielder (P: pedestrian \\or V: vehicle}}    & \multirow{2}{*}{2682} & \multirow{2}{*}{1.29} \\ \\
    \midrule
    \bf Independent variables  \\ (Continuous)  & \bf Mean &\bf Std Dev \\
    \midrule
    Vehicle speed  & 15.93 & \multicolumn{1}{c}{10.45} \\
    
   Number of pedestrians & 186.33 & \multicolumn{1}{c}{185.41} \\
   Number of lanes at main street &2.41 & 0.78\\
   Crossing width (major) &49.75 &4.88 \\
   Major AADT & 10220.47 & 4047.36 \\
   Minor AADT & 1350 &355.98 \\
Distance to the nearest park &0.25 &0.12 \\
Distance to the nearest school & 0.51 &0.42 \\
    Number of bus stops& 2.39   & 2.51 \\

    \midrule
    \textbf{\bf Independent variables} \\ (Categorical)  & \textbf{Categories} & \textbf{Description} \\
    \midrule
   Close call & 2 & Yes, No \\
   Pedestrian origin & 4 & \makecell{A, B, C, D,  4 intersection corners} \\
    Pedestrian destination & 4  & \makecell{A, B, C, D,  4 intersection corners} \\
    Pedestrian stop midway & 2 & Yes, No \\

    Type of yielding vehicle & 5     & ``Car", ``SUV",``Utility vehicle",``Truck",``Bus" \\
    Opposite direction yield & 2 & Yes, No \\
    Following vehicle & 2 & Yes, No \\
    Posted speed limit & 2 & 30, 35 \\
    Bike lanes &2 & Yes, No \\
    Weather & 3 & No precipitations, raining, snowing \\
    Signage & 2 & No signs, Signed \\
    Markings & 3 & Unmarked, Continental marking, Standard marking \\
    Presence of single family within 1 block  & 2 & Yes, No\\
    Presence of apartments within 1 block   & 2 & Yes, No\\
    Presence of commercial within 1 block & 2 & Yes, No\\
    \makecell{Presence of Gas/Station\\Convenient Store within 1 block } & 2 & Yes, No\\
    Presence of Restaurants/Bars within 1 block  & 2 & Yes, No\\
    Presence of parking lots within 1 block  & 2 & Yes, No\\
    Street Lighting & 2 & Yes, No \\
    Road surface & 2 & Dry, Wet \\
    Presence of on-street parking &2 &Yes, No \\
    PAWS Score &19 & \makecell{Priority Areas for \\Walking Study scores for half-mile hexagons \\based on equity, safety, health, \\infrastructure, and land use factors}\\
    Tree cover & 5& The number of crossing points covered by tree \\
    
Interaction type & 5     & A, B, C, D, E \\
 Pedestrian type & 8     & A, B, C, D, E, F \\

    \bottomrule
    \end{tabular}}
  \label{tab:Variables_description}%
\end{table}%

\end{document}